\documentclass[journal]{IEEEtran}
\usepackage[ruled,linesnumbered,vlined]{algorithm2e}
\usepackage{hyperref}
\usepackage{amsmath,amsfonts}
\usepackage{graphicx}
\usepackage{algpseudocode}
\usepackage{enumitem}
\usepackage{booktabs}
\usepackage{tabularx}
\usepackage{amsthm}
\theoremstyle{remark}
\newtheorem{remark}{\indent Remark}

\newtheorem{theorem}{\indent Theorem}

\newtheorem*{prf}{\indent Proof}

\usepackage{multirow}
\usepackage{array}
\usepackage{pifont}
\usepackage{algpseudocode}
\usepackage[normalem]{ulem}
\usepackage{multirow}
\usepackage{caption}
\usepackage{siunitx}
\usepackage{booktabs}
\usepackage[utf8]{inputenc} 
\usepackage{multirow}
\usepackage{newunicodechar}
\newunicodechar{✓}{\checkmark} 
\usepackage{changepage}
\usepackage{subcaption}
\usepackage{afterpage}
\usepackage{array}
\usepackage{mathrsfs}
\usepackage{bm}
\usepackage{url}
\usepackage[switch]{lineno}
\usepackage[table]{xcolor}   
\usepackage{amsmath,amsfonts}
\usepackage{algpseudocode}
\usepackage{xcolor}
\usepackage{xpatch}

\usepackage{siunitx}
\usepackage{siunitx}
\definecolor{darkred}{RGB}{192,0,0} 

\usepackage[
backend=biber,
style=ieee,       
sorting=none         
]{biblatex}

\hypersetup{
    colorlinks=true,     
    linkcolor=blue,   
    urlcolor=magenta,
    citecolor=blue 
}

\begin{document}
\title{Learning Self-Growth Maps for Fast and Accurate Imbalanced Streaming Data Clustering}

\author{Yiqun~Zhang,~\IEEEmembership{Senior Member,~IEEE},
        Sen Feng,
        Pengkai Wang,
        Zexi Tan,
        Xiaopeng Luo,\\
        Yuzhu Ji~\IEEEmembership{Member,~IEEE},
        Rong Zou,
        and Yiu-ming Cheung,~\IEEEmembership{Fellow,~IEEE}

\thanks{Yiqun Zhang, Sen Feng, Pengkai Wang, Zexi Tan, Xiaopeng Luo, and Yuzhu Ji are with the School of Computer Science and Technology, Guangdong University of Technology, Guangzhou, China (e-mail: yqzhang@gdut.edu.cn, \{2112305084, 2112405278, 3123004194\}@mail2.gdut.edu.cn, gordonlok@foxmail.com, yuzhu.ji@gdut.edu.cn).}
\thanks{Rong Zou and Yiu-ming Cheung are with the Department of Computer Science, Hong Kong Baptist University, Hong Kong, SAR, China (e-mail: \{rongzou, ymc\}@comp.hkbu.edu.hk).}
\thanks{Corresponding author: Yuzhu Ji and Yiu-ming Cheung.}
}

\markboth{IEEE Transactions on Neural Networks and Learning Systems}{ }

\maketitle

\begin{abstract}
Streaming data clustering is a popular research topic in data mining and machine learning. Since streaming data is usually analyzed in data chunks, it is more susceptible to encounter the dynamic cluster imbalance issue. That is, the imbalance ratio of clusters changes over time, which can easily lead to fluctuations in either the accuracy or the efficiency of streaming data clustering. Therefore, we propose an accurate and efficient streaming data clustering approach to adapt the drifting and imbalanced cluster distributions. We first design a Self-Growth Map (SGM) that can automatically arrange neurons on demand according to local distribution, and thus achieve fast and incremental adaptation to the streaming distributions. Since SGM allocates an excess number of density-sensitive neurons to describe the global distribution, it can avoid missing small clusters among imbalanced distributions. We also propose a fast hierarchical merging strategy to combine the neurons that break up the relatively large clusters. It exploits the maintained SGM to quickly retrieve the intra-cluster distribution pairs for merging, which circumvents the most laborious global searching. It turns out that the proposed SGM can incrementally adapt to the distributions of new chunks, and the \underline{S}elf-gr\underline{O}wth map-guided \underline{H}ierarchical merging for \underline{I}mbalanced data clustering (SOHI) approach can quickly explore a true number of imbalanced clusters. Extensive experiments demonstrate that SOHI can efficiently and accurately explore cluster distributions for streaming data.
\end{abstract}

\begin{IEEEkeywords}
Cluster analysis, streaming data, imbalanced data, self-organizing map, drift adaptation, efficient algorithms.
\end{IEEEkeywords}

\IEEEpeerreviewmaketitle

\section{Introduction}\label{section1}

\begin{figure}[!t]
    \centering
    \resizebox{1.0\linewidth}{!}{\includegraphics{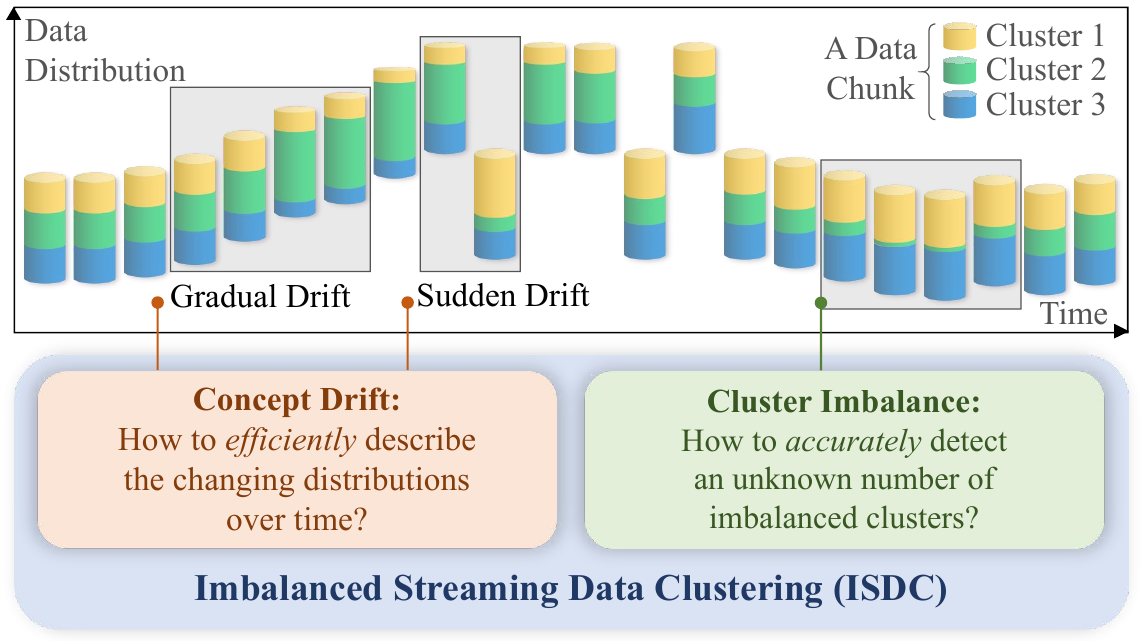}}
    \caption{Key problems in imbalanced streaming data clustering.}
    \label{fig:ISDC}
\end{figure}

\IEEEPARstart{S}{treaming} data, specifically the datasets with flowing-in or updates of its objects over time, is prevalent in various fields, such as market research, health big data analysis, and the internet of things \cite{gaber2005mining, TNNLS2, zhao2022heterogeneous}. Due to the lack of readily available labels for streaming data, clustering that gathers similar data objects into a certain number of groups becomes indispensable for data analysis. Cluster analysis of streaming data is often conducted on a chunk-by-chunk basis to ensure sufficient statistical information \cite{Alaettin2021Data_stream}. However, the non-uniform co-occurrence of data objects from different distributions at the same time, along with shifts in data distributions over time \cite{TNNLS4}, will frequently lead to the emergence of imbalanced clusters, i.e., clusters with very different numbers of objects. Fig.~\ref{fig:ISDC} shows the causes of such a problem, which puts forward a new clustering challenge, i.e., how to accurately explore and efficiently adapt to the imbalanced cluster distributions of streaming data chunks.

Imbalanced streaming data introduces more complex and challenging issues to clustering, which are manifested into two key aspects: 1) Uncertainty in the number of clusters, and 2) Laborious detection of imbalanced clusters. On the one hand, due to the very different and changing sizes of clusters, it is difficult to dynamically determine an appropriate number of clusters, leading to unsatisfactory clustering accuracy. On the other hand, existing imbalanced data clustering solutions \cite{lu2019self} usually involve two phases: i) partition all $n$ objects into $m$-scale microclusters ($m \gg k^*$, where $k^*$ is the true number of clusters), and ii) perform $m^2$-scale merging of the microclusters to form final clusters. Such a time-consuming $O(m^2)$-complex process prevents the existing imbalanced data clustering approaches from efficient streaming data analysis, while the fast clustering techniques struggle in exploring imbalanced clusters due to the overlook of relatively small clusters. Subsequently, we analyze existing relevant clustering solutions from the perspective of Imbalanced Streaming Data Clustering (ISDC).

Partitional clustering algorithms demonstrate a certain level of capability in addressing ISDC. In this stream, the most classic method would be the $k$-means \cite{kmeans} algorithm. Thanks to its simplicity and efficiency, it can be applied to the analysis of streaming data with linear complexity. However, both $k$-means and most of its variants (e.g., \cite{FSWL, RPCL}) tend to produce balanced clusters \cite{Abiodun2023CR}. To specifically address the imbalance issue, the Multi-Center (MC) clustering algorithm \cite{liang} has been proposed to partition the whole dataset into many small subclusters, and then successively merge the closer pairs to avoid neglecting relatively small clusters. As its performance is very sensitive to initialization, a self-adaptive clustering method called SMCL \cite{lu2019self} has been proposed, which more robustly generates a proper number of seed points through competitive learning \cite{RPCCL, Xu2019DenSIONN, Jia2014CPCL} to describe the data distribution for exploring imbalanced clusters. 
A novel rough set-based approach M3W \cite{m3w} has also been proposed to generate a hierarchical data distribution structure that informatively facilitates the formation of small subclusters with clear boundaries. However, similar to MC and SMCL, it involves relatively laborious computation due to the $m^2$-scale search of many ($m$-scale) seed points, making it unsuitable for streaming data.

Density-based clustering methods determine object-cluster affiliation according to the local density of objects \cite{liu2018shared}, and they demonstrate a certain degree of capability in handling imbalanced data. For instance, Density Peaks Clustering (DPC) \cite{DPC} identifies data objects with relatively higher local density as cluster centers, and can thus explore smaller clusters with relatively prominent local density. To achieve more reasonable density quantification, an algorithm called FKNN-DPC \cite{Xie} has been proposed, employing fuzzy weighted $k$-nearest neighbor as a density measure. With a more appropriate density measure, FKNN-DPC achieves better performance in detecting imbalanced clusters in comparison with the original DPC. Later, an approach called Local Density Peaks for Imbalanced data (LDPI) \cite{Tong_Wang_Liu} has been proposed, adopting an adaptive subcluster construction scheme, which forms more subclusters than true clusters to enhance the detection of imbalanced clusters. However, LDPI has an object-wise quadratic time complexity, thus outside the consideration of efficient ISDC.

To specifically achieve efficient cluster analysis, fast clustering solutions have emerged. One of the most conventional methods is StreamKM++ \cite{StreamKM++}, which integrates a software acceleration architecture and $k$-means++ \cite{k-means++} to dynamically update cluster centers to incrementally fit streaming data distribution. The efficiency of density-based clustering has also been improved by adapting DPC to streaming data, which demonstrates superior clustering performance compared to other streaming data clustering methods \cite{liu2018shared}. However, dynamic DPC is not robust to hyper-parameters when dealing with non-stationary streaming data. To further address this issue, AMD-DPC \cite{AMD-DPC} was proposed, which adopts a graph-based data structure for local density updates. This algorithm saves computation cost while remaining accurate, showing potential for real-time big data processing. By inheriting the efficiency of partitional clustering and the unbiased cluster discovery capability of density-based clustering, a fast and accurate clustering algorithm called IGMTT \cite{Cheung2019DenIGMTT} has been presented. However, the above-mentioned fast algorithms have yet to consider imbalanced clusters, and may thus yield unsatisfactory clustering results in the ISDC tasks.

To the best of our knowledge, most conventional clustering assumes that the true number of clusters $k^*$ is given by the users based on prior knowledge of the data \cite{Alaettin2021Data_stream, Abiodun2023CR}. However, for ISDC, where the distribution of the current new data chunks may continue to change, it is difficult to obtain prior knowledge about $k^*$ in advance. Combining all the above analysis, it can be concluded that existing methods only consider one of the factors of imbalance or streaming, and most of them do not take into account the case of unknown $k^*$. Therefore, there is an urgent need to achieve fast and accurate ISDC without knowing $k^*$.

This paper, therefore, proposes a fast and accurate approach called \underline{\textbf{S}}elf-gr\underline{\textbf{O}}wth maps-guided \underline{\textbf{H}}ierarchical merging for \underline{\textbf{I}}mbalanced data clustering (SOHI, pronounced `so high!') for ISDC. First, to realize an efficient distribution description for data chunks, we propose a Self-Growth Map (SGM) learning algorithm that can quickly and incrementally adapt the distribution via connected neurons generated in need. The maps use a 3-neuron triangle as the basic geometry unit for growth. Compared to the self-organizing map that adopts pre-sized grids, our SGM, which grows in a triangular manner, enables flexible distribution exploration and can incrementally adapt to the changing distributions. Based on the distribution described by the abundant map neurons, relatively small clusters can be effectively captured. Subsequently, further exploration of the imbalanced clusters is performed by hierarchically merging the adjacent data objects corresponding to the neurons. The topological structure of SGM is fully utilized to accelerate the laborious hierarchical merging by serving as a retrieval structure, allowing only closely connected neurons to merge. According to our analysis, such a map retrieval-based acceleration considerably improves the time complexity. In comparison with the state-of-the-art counterparts, SOHI demonstrates its superiority in ISDC as it sufficiently improves time complexity, while still being competitive in clustering accuracy. Comprehensive experiments have been conducted to illustrate its promising performance. The main contributions are summarized into four-fold: 
\begin{enumerate}
\item A new paradigm called SOHI is proposed for ISDC. The distribution information provided by the learned SGM is thoroughly exploited to relieve the trade-off between efficiency and accuracy in ISDC.
\item We propose self-growing maps named SGM with triangles as the basic growth unit. Such a design avoids the missing of small clusters, and can flexibly adapt to streaming chunks incrementally.
\item To ensure an efficient imbalanced cluster detection, a fast hierarchical merging mechanism is designed to fully utilize the similarity of local distributions reflected by the SGM, thus considerably avoiding meaningless searches.
\item An imbalanced streaming data chunk generator is presented to simultaneously simulate the changing of cluster number and cluster size in real ISDC scenarios. It ensures a more convincing evaluation and can be a universal experimental tool in studying ISDC.
\end{enumerate}

The rest of this paper is organized as follows: Section \ref{section2} reviews the related work. Section \ref{section3} presents the problem statement and preliminaries, and Section \ref{section4} introduces the proposed SOHI with time complexity analysis. Section \ref{section5} showcases experimental results with in-depth observations. Finally, we conclude the whole work in Section \ref{section6}. 

\section{Related Work}\label{section2}

This section overviews the research topics relevant to this paper, including streaming data clustering, imbalanced data clustering, and distribution learning techniques.

\subsection{Streaming Data Clustering}

Data stream clustering partitions a series of data chunks into compact clusters, which places higher requirements on efficiency than static data clustering. Therefore, the streaming data clustering method strives to strike a balance between accuracy and efficiency \cite{Alaettin2021Data_stream}. Many of these algorithms are variations of traditional clustering methods like the partitional $k$-means \cite{Abiodun2023CR} (e.g., \cite{Xu2019DenSIONN,StreamKM++,BIRCH,clustream}), density-based DBSCAN (e.g.,\cite{DP2,DP1,peng2025weighted}), and hierarchical clustering \cite{ODAC}. Based on the brief introduction in Section~\ref{section1}, we know that they either introduce approximation-based acceleration from the computational level, such as StreamKM++ \cite{StreamKM++}, or replace the original modules involving laborious computation with more efficient alternatives from the algorithmic level, e.g., IGMTT \cite{Cheung2019DenIGMTT}. However, they inevitably introduce hyper-parameters that are difficult to tune and may cause unstable performance.

Since constructing and merging microclusters by measuring the Euclidean distances between high-dimensional data objects is challenging, OSRC \cite{Chen2025OnlineSparse} utilizes low-dimensional projection to effectively select an appropriate number of representative data objects, but may be sensitive to the selection of hyperparameters. To circumvent this, the work proposed in \cite{zhang2017} uses the Davies-Bouldin Index (DBI) to guide the optimization of clustering. However, its scalability is limited by the DBI computation based on static data. Accordingly, the work \cite{moshtaghi2019} develops incremental Xie-Beni (XB) and DBI indices to monitor the streaming clustering process of $k$-means type algorithms. To extend the above methods to process multi-view data, a multi-view support vector domain description model \cite{TNNLS1} has been proposed to capture cluster evolution and discover arbitrarily shaped clusters with limited computing resources. Nevertheless, streaming data clustering remains a challenging issue due to the unavoidable trade-off between efficiency and accuracy.

\subsection{Imbalanced Data Clustering}

Imbalanced data clustering, where the scale varies for different clusters, has attracted much more attention in real data mining applications \cite{Imbalancedfirst}. In addition to the density-based clustering methods introduced in Section~\ref{section1}, which naturally has a certain ability to handle imbalanced clusters, using many prototypes to capture the micro distributions is also considered one of the most effective ways to specifically avoid missing small clusters. Such a principle has been commonly adopted by the works \cite{liang,k-means1,zhang2025adaptive,Shenghong-2024,zhang2024towards}. The undersampling strategy is also one of the solutions for imbalanced clustering, but it often faces the difficulties of gradient explosion and insufficient learning experience of positive data objects. Recently, \cite{Huang2024PreferenceSampling} proposes an informative undersampling and boundary expansion strategy to deal with it. However, they typically use a pre-defined number of prototypes, which makes the clustering performance highly sensitive to different datasets with various distributions. Hence, their recent variants propose to adaptively generate prototypes for distribution description.

A method called SMCL \cite{lu2019self} achieves incremental prototype learning by gradually adding seed points driven by a competitive learning mechanism to prevent the issue of dead units \cite{RPCCL}. However, this approach lacks robustness to noise and is extremely computationally expensive due to the recursive seed points generation and merging. To address this, LDPI \cite{Tong_Wang_Liu} designs an initial subcluster generation scheme, improving the clustering method of DPC \cite{DPC} by automatically identifying noise points and initial subcluster centers. According to the nearest-neighbor principle, the remaining objects are classified as subcluster centers to represent local micro distributions. MCNS \cite{Li_Zhou_Zeng_Chan_2023} further introduces a measure based on the reconstruction rate to select the appropriate number of clusters, enhancing convergence speed while ensuring accuracy. However, all the above-mentioned methods seek the optimal solution through iterative searching on $n$-scale prototypes, resulting in quadratic-level time complexity.

\subsection{Distribution Learning}
Common unsupervised distribution learning approaches include: 1) representation learning \cite{chen2024qgrl,zhang2025learning,feng2025robust} that learns to project the data objects from the original distance space into a more cluster-discriminative space; 2) data summarization \cite{Datasum1, Datasum2} that uses a set of prototypes to describe the data distribution; and 3) Self-Organizing Map (SOM) \cite{Kohonen_1990, J2000som, luo2024efficient} that trains a low-dimensional map to simultaneously realize dimensionality reduction and distribution description on datasets. Since the latter two types are more efficiency-promising under the scenario of ISDC, we further discuss them below.

Among the summarization-based methods, data bubbles \cite{Data-Bubbles} and its variants \cite{Fast-Hierarchical, Samer-Nassar} summarize data distribution by randomly initializing a set of prototypes to incorporate nearby data objects into groups (i.e., data bubbles). In general, their performance is sensitive to the compression rate and the initialization of prototypes. Later, the work in \cite{QualityZhang2016} relieves the sensitivity issue by specifically training prototypes to fit the distribution. To further preserve the embedded hierarchies of more complex data distribution, hierarchical summarization approaches \cite{Cheung2019DenIGMTT, BIRCH, Shenghong-2024} have been proposed to partition the data into microclusters, and then construct dendrograms for multi-granular distribution summarization.

SOM \cite{Kohonen_1990} that trains neurons constrained by mesh topology in a competitive learning way has been proposed to summarize high-dimensional data into a low-dimensional map for unbiased any-shape cluster detection \cite{J2000som, Jose2001Kmam}. To improve the adaptability of SOM to represent more complex data distributions, more advanced SOMs \cite{ASOM, RALSOM, SOFM} that introduce asymmetric connection weights, dynamic learning rates, and adaptive neighborhood relationships, respectively, have been proposed in recent years. Given that conventional SOMs initialize fix-sized and grid-connected neurons, they are inflexible in distribution exploration of streaming data. To address this issue, growing self-organizing maps \cite{Bernd1999GSO,Cheung2019DenIGMTT} have been developed to generate neurons on demand to fit data distributions. The growing hierarchical map \cite{GHSOM} has been proposed to recursively grow and refine the coarse-grained neurons to represent complex data distributions. Most recently, the method proposed in \cite{David2022PBKM} further generalizes SOM to a sparse dictionary of prototypes with flexible free-to-update/remove neighboring relationships, thus facilitating efficient and accurate online streaming data classification. However, the above-mentioned approaches either operate in a supervised scenario, or do not specifically address the imbalance of distributions, preventing them from tackling the ISDC problem. Furthermore, SOM has also been incorporated into deep learning framework \cite{TNNLS3}. Although it facilitates a convenient end-to-end clustering, the data scale requirement for model training and the inability to adapt to concept drift limit its usage in ISDC.

\section{Problem Statement}\label{section3}

The primary cause of the unsatisfactory streaming clustering performance is the difficulty in exploring imbalanced clusters, known as the Imbalanced Streaming Data Clustering (ISDC) problem demonstrated in Fig.~\ref{fig:ISDC}. Assuming data objects are continuously generated or collected, which can be more formally denoted as a data stream $X_\Gamma=\left\{ X^\iota \right\}_{\iota=1}^{N}$ comprising $N$ data chunks $X^\iota\in \mathbb{R}^{n\times d}$ formed at different time-stamps $\iota$. Generally, data streams are considered to be unbounded (i.e., $N\rightarrow \infty$). Each data chunk consists of $n$ objects denoted as a collection of $n$ vectors $X^\iota=\left\{ \mathbf{x}_{1}^{\iota},\mathbf{x}_{2}^{\iota},\ldots,\mathbf{x}_{n}^{\iota} \right\}$, where $\mathbf{x}_{j}^{\iota}\in \mathbb{R}^d$ represents the $j$-th data object within $X^\iota$. 

For ISDC, assume the objects of a chunk $X^\iota$ can be distributed to $k^\iota$ clusters denoted as $C^{\iota}=\{C^{\iota}_1,C^{\iota}_2,\ldots, C^{\iota}_{k^\iota}\}$ with the corresponding cluster centers $S^{\iota}=\{\mathbf{s}^\iota_1, \mathbf{s}^\iota_2, \ldots, \mathbf{s}^\iota_{k^\iota}\}$. A $j$-th cluster $C^{\iota}_j$ is a subset of $X^\iota$. 
The conventional cluster objective is to partition $X^\iota$ into $C^{\iota}$ that minimizes:
\begin{equation}
     SSQ( X^\iota,S^\iota ) = \sum_{j=1}^{k^\iota}{\sum_{\mathbf{x}\in C^\iota_j}{\| \mathbf{x} - \mathbf{s}^\iota_j \|_2}},
\end{equation}
which is the sum of squared distances~\cite{SSD}, and $\|\cdot\|_2$ denotes the L-2 norm. 

When data object generation is biased towards certain clusters, the cluster-imbalanced data chunks arise, where the number of objects in the relatively large cluster can significantly exceed that in the relatively small one, which can be reflected by an Imbalance Ratio (IR):
\begin{equation}\label{equal2}
  IR=\frac{\max(\text{card}(C_1^\iota), \text{card}(C_2^\iota), ...,\text{card}(C_{k^\iota}^\iota))}{\min(\text{card}(C_1^\iota), \text{card}(C_2^\iota), ...,\text{card}(C_{k^\iota}^\iota))},
\end{equation}
where $\text{card}(\cdot)$ is a function that counts the number of elements of a set, and $IR$ is actually the ratio between the sizes of the largest and the smallest clusters. Intuitively, $IR=1$ indicates an extremely balanced case, while larger $IR$ indicates a more severe cluster imbalance. Since most existing clustering algorithms implicitly assume balanced clusters and perform static data clustering, how to timely capture the key smaller clusters in ISDC is the core problem to be tackled.

\begin{figure}[!t]
    \centering
    \resizebox{1.0\linewidth}{!}{\includegraphics{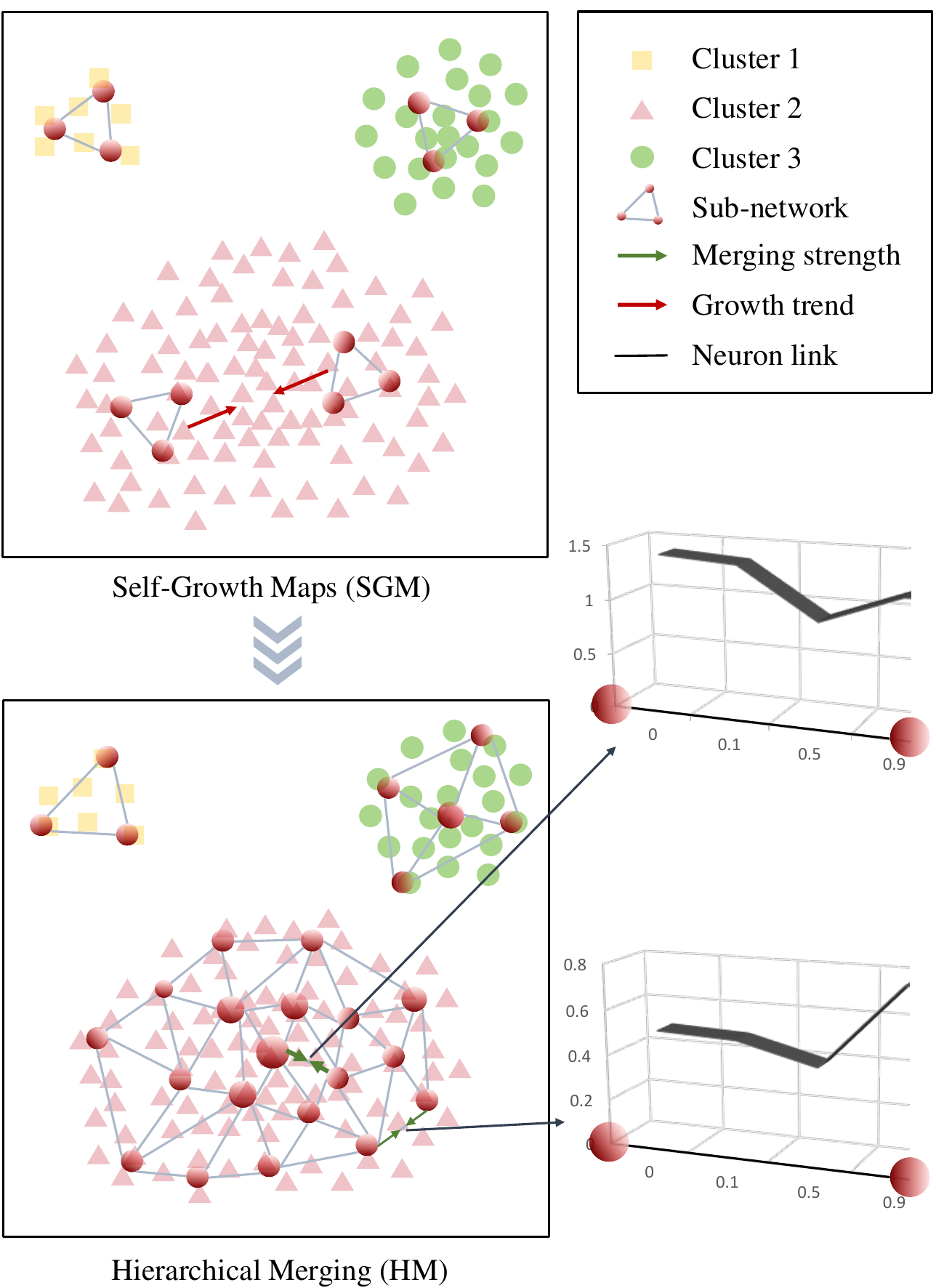}}
    \caption{Overview of the proposed SOHI framework.}
    \label{SOHI}
\end{figure}

\section{Proposed Method}\label{section4}

This section provides a detailed description of the proposed SOHI approach for ISDC. SOHI contains two main steps: 1) Adaptive chunk distribution representation by Self-Growth Maps (SGM), and 2) Hierarchical Merging (HM) for imbalanced cluster exploration. The overview of the SOHI approach is shown in Fig.~\ref{SOHI}. SGM first initializes multiple subnetworks and lets them grow to fit the local object distribution (upper left in Fig.~\ref{SOHI}). When the overall data distribution can be sufficiently represented by the SGM, the neurons are hierarchically merged by the HM (lower left in Fig.~\ref{SOHI}) to explore an optimal number of imbalanced clusters based on the density gaps between merged neurons (lower right in Fig.~\ref{SOHI}).

\subsection{SGM: Self-Growth Maps}\label{section4A}

For ISDC, it is crucial to achieve fast microcluster exploration for distribution description. Most conventional SOM-based methods initialize and train a complete map to represent the data distribution through its neurons, where excessive neurons can appear between adjacent clusters, making them indistinguishable. More specifically, the neurons are usually treated as microcluster centers and will guide the merging of microclusters to form prominent clusters. If redundant neurons between adjacent clusters are created, clusters might be mistakenly merged, severely degrading the clustering performance. Growing Cell Structures (GCS) \cite{GCSOM}, an incremental SOM, is promising as it dynamically adapts to the changing distributions of streaming data. However, it attempts to maintain a complete map with all neurons connected, which may still incur the redundant neuron effect. Although it provides a threshold-based strategy for redundant neuron elimination, the threshold tuning is still non-trivial.

Inspired by GCS, the Self-Growth Map (SGM) is proposed to train multiple subnetworks for rapid data distribution fitting. A network $A$ consisting of $T$ subnetworks is initialized:
\begin{equation}
\label{n3_1}
    A = \bigcup\limits_{l=1}^{T} A_l,
\end{equation}
where $T$ is typically set at a value larger than the optimal number of clusters $k^*$ to avoid missing relatively small clusters. To ensure that each subnetwork can appropriately fit the local data distribution without overlapping, rapid Poisson disk sampling is employed for generating initial subnetworks \cite{Bridson_2007} (each of which is a basic triangle 3-neuron network), which are allowed to grow independently. Planar triangle structures are adopted by the growing network rather than the high-dimensional hypertetrahedra as in the original GCS for an efficient purpose. Given a chunk $X^\iota$, subnetworks, e.g., $A_l$, are trained to fit the local object distribution by:
\begin{equation}
\label{n7}
\mathbf{z}_{l,s}^{\text{new}} = \mathbf{z}_{l,s}^{\text{old}} + \epsilon_b (\mathbf{x}_{j}^{\iota} - \mathbf{z}_{l,s}^{\text{old}}),\ 
\end{equation}
if $\mathbf{z}_{l,s}$ is the Best Matching Neuron (BMN) of object $\mathbf{x}^\iota_j$ determined by: 
\begin{equation}
\label{n4}
\mathbf{z}_{l,s}=\underset{\mathbf{z}_{i,r} \in A}{\mathrm{argmin}} \| \mathbf{x}_{j}^{\iota} - \mathbf{z}_{i,r} \|_2.
\end{equation}
Note that $\mathbf{z}_{l,s}$ is a $d$-dimensional vector representing the $s$-th neuron of $A_l$, as the training of neurons is driven by the $d$-dimensional data objects from $X^\iota$. 
Moreover, to ensure a smooth and efficient update of the BMNs, a small learning rate $\epsilon_b$ is adopted in Eq.~(\ref{n7}). To ensure a structural update of the subnetwork, the update provided by $\mathbf{x}_{j}^{\iota}$ is also propagated to the neurons that are directly connected with the BMN: 
\begin{equation}
\label{n7_1}
    \mathbf{z}_{l,c}^{\text{new}} = \mathbf{z}_{l,c}^{\text{old}} + \epsilon_{\Omega} (\mathbf{x}_{j}^{\iota} - \mathbf{z}_{l,c}^{\text{old}}),
\end{equation}
$$\text{s.t.}\ \mathbf{z}_{l,c} \in \Omega(\mathbf{z}_{l,s}),$$
where $\Omega(\mathbf{z}_{l,s})$ represents the set of 1-hop adjacent neurons of $\mathbf{z}_{l,s}$, and $\epsilon_{\Omega}$ is the corresponding learning rate. Since only the BMN and its 1-hop neighbors are trained w.r.t. each object $\mathbf{x}_{j}^{\iota}$, an efficient adaptation to new data distributions can be achieved without involving all the subnetwork's neurons. By randomly selecting data objects from $X^\iota$ for the above training, the network gradually fits the distribution of $X^\iota$.

Due to the distribution complexity, mapping too many distant data objects to a neuron will lead to an improper distribution representation. Such an effect can be reflected by the accumulated BMN distance, which we call BMN inadaptability. For each training input $\mathbf{x}_{j}^{\iota}$, its corresponding BMN inadaptability is updated by: 
\begin{equation}
\label{n8}
\tau^{\text{new}}_{\mathbf{z}_{l,s}} = \tau^{\text{old}}_{\mathbf{z}_{l,s}} + \|\mathbf{x}_{j}^{\iota} - \mathbf{z}_{l,s}\|_2.
\end{equation}
In the same subnetwork $A_l$, the neurons other than the BMN typically exhibit lower inadaptability with respect to the current input. Therefore, their inadaptability values are slightly decreased, with the reduction rate governed by a parameter $\alpha$:
\begin{equation}
\label{n9}
\tau^{\text{new}}_{\mathbf{z}_{l,q}} = \tau^{\text{old}}_{\mathbf{z}_{l,q}} -\alpha\tau^{\text{old}}_{\mathbf{z}_{l,q}},
\end{equation}
$$\text{s.t.}\  {\mathbf{z}_{l,q}} \neq \mathbf{z}_{l,s}.$$

When $A$ has been trained to adapt to $\rho$ data objects, it is necessary to evaluate whether the network needs to create new neurons or merge nearby subnetworks. A reasonable evaluation method is to compare the distance between the neurons with the highest and second highest inadaptability, e.g., $\mathbf{z}_{i,v}$ and $\mathbf{z}_{j,g}$, which belong to two different subnetworks, given by:
\begin{equation}
\mathbf{z}_{i,v} = \underset{\mathbf{z}_{i,s}\in A}{\mathrm{argmax}}\tau_{\mathbf{z}_{i,s}}\ \ \text{and}\ \ 
\mathbf{z}_{j,g} = \underset{\mathbf{z}_{j,s}\in A \setminus \{A_i\}}{\mathrm{argmax}}\tau_{\mathbf{z}_{j,s}}. \label{eq:zi_j}
\end{equation}
If the distance between $\mathbf{z}_{i,v}$ and $\mathbf{z}_{j,g}$ is larger than that between $\mathbf{z}_{i,v}$ and its furthest adjacent neuron $\mathbf{z}_{i,q}$, i.e., 
\begin{equation}
\label{add_neurs}
\| \mathbf{z}_{i,v} - \mathbf{z}_{j,g} \|_2 > \| \mathbf{z}_{i,v} - \mathbf{z}_{i,q} \|_2, 
\end{equation} 
$$\text{s.t.}\  \mathbf{z}_{i,q} = \underset{\mathbf{z}_{i,g} \in \Omega({\mathbf{z}_{i,v}})}{\mathrm{argmax}} \| \mathbf{z}_{i,g} - \mathbf{z}_{i,v} \|_2,$$
then it suggests that $\mathbf{z}_{i,v}$ may suffer from under-adaptation and be insufficient to properly represent its associated objects. To address this, a new neuron $\mathbf{z}_{i,h} = (\mathbf{z}_{i,v} + \mathbf{z}_{i,q}) / 2$ is created and connected to the common neighbors of $\mathbf{z}_{i,v}$ and $\mathbf{z}_{i,q}$, preserving the basic triangle topology of the network.

To facilitate sustainable map growth, an inadaptability value $\tau^{\text{new}}_{\mathbf{z}_{i,h}}$ should be assigned to the new neuron $\mathbf{z}_{i,h}$:
\begin{equation}
\label{n15}
\tau^{\text{new}}_{\mathbf{z}_{i,h}} = \sum_{\mathbf{z}_{i,c} \in \Omega(\mathbf{z}_{i,h})} \frac{{\zeta}^{\text{old}}_{\mathbf{z}_{i,c}} - {\zeta}^{\text{new}}_{\mathbf{z}_{i,c}}}{{\zeta}^{\text{old}}_{\mathbf{z}_{i,c}}} \tau^{\text{old}}_{\mathbf{z}_{i,c}},
\end{equation}
which is the sum of the reduced inadaptability of its neighboring neurons. The reduced amount of inadaptability is reflected by the receptive field of a neuron:
\begin{equation}
\label{n5}
\zeta_{\mathbf{z}_{i,c}} = \frac{1}{ \text{card}(\Omega(\mathbf{z}_{i,c})) } \sum_{\mathbf{z}_{i,t} \in \Omega(\mathbf{z}_{i,c})} \lVert \mathbf{z}_{i,c} - \mathbf{z}_{i,t} \rVert_2,
\end{equation}
which is the average distance between $\mathbf{z}_{i,c}$ and its neighboring neurons. As the new $\mathbf{z}_{i,h}$ splits up the receptive fields from its neighboring neurons, the inadaptability of each neighboring neuron $\mathbf{z}_{i,c}$ is updated with a corresponding reduction as:
\begin{equation}
\label{n14}
\tau^{\text{new}}_{\mathbf{z}_{i,c}} = \tau^{\text{old}}_{\mathbf{z}_{i,c}} - \frac{{\zeta}^{\text{old}}_{\mathbf{z}_{i,c}} - {\zeta}^{\text{new}}_{\mathbf{z}_{i,c}}}{{\zeta}^{\text{old}}_{\mathbf{z}_{i,c}}} \tau^{\text{old}}_{\mathbf{z}_{i,c}}, 
\end{equation}
$$\text{s.t.}\  \quad \mathbf{z}_{i,c} \in \Omega(\mathbf{z}_{i,h}).$$

If the distance between $\mathbf{z}_{i,v}$ and $\mathbf{z}_{j,g}$ is not larger than the distance between $\mathbf{z}_{i,v}$ and $\mathbf{z}_{i,q}$, it implies that the two corresponding subnetworks $A_i$ and $A_j$ are too close, and a merging procedure should be launched. Specifically, an edge is created to connect $\mathbf{z}_{i,v}$ and $\mathbf{z}_{i,q}$ to $\mathbf{z}_{j,g}$, maintaining the triangular topology within and across subnetworks. The complete algorithm of inadaptability-guided neuron creation and subnetwork merging is summarized in Algorithm~\ref{Alg_IGCM}.

\begin{algorithm}[!t]
\caption{IGCM: Inadaptability measure-Guided neuron Creation or subnetwork Merging}
\label{Alg_IGCM}
\SetAlgoLined
\KwIn{Network $A$.}
\KwOut{Updated network $A$ with created neurons or merged subnetworks.} 
Find $\mathbf{z}_{i,v}$ and $\mathbf{z}_{j,g}$ with the largest and second largest inadaptability by Eq.~(\ref{eq:zi_j})\;
Find the furthest adjacent neuron $\mathbf{z}_{i,q}$ of $\mathbf{z}_{i,v}$ from $A_i$\;
\uIf{$\| \mathbf{z}_{i,v} - \mathbf{z}_{j,g} \|_2 > \| \mathbf{z}_{i,v} - \mathbf{z}_{i,q} \|_2$}{
    Create $\mathbf{z}_{i,h}$ with its inadaptability $\tau_{\mathbf{z}_{i,h}}$ by Eq.~(\ref{n15})\; 
    Update inadaptabilities of $\Omega(\mathbf{z}_{i,h})$ by Eq.~(\ref{n14})\; 
    Create edges to connect $\mathbf{z}_{i,h}$ to $\mathbf{z}_{i,v}$ and $\mathbf{z}_{i,q}$\;
    }
\Else{
   Connect $\mathbf{z}_{i,v}$ and $\mathbf{z}_{i,q}$ to $\mathbf{z}_{j,g}$, merge $A_i$ and $A_j$\;
    }
\end{algorithm}

\begin{remark}\label{rmk:Merits of Subnetworks}
\textbf{Merits of Subnetworks:} The multi-subnetwork design naturally circumvents the thorny elimination of redundant neurons located at the distribution boundary of clusters. Moreover, it allows for the flexible removal of subnetworks that do not fit the data distribution in a new data chunk without affecting the structure of the other subnetworks. In addition, parallel computing can be performed to accelerate the training of the independent subnetworks before their merging.
\end{remark}

\begin{remark}\label{subnework Merging}
\textbf{Rationality of the IGCM algorithm:} 
Subnetwork merging is conducted by considering the neurons with high inadaptability and their inter-subnetwork distance. Inadaptability filters most neurons that can well represent the corresponding objects, 
and the designed merging process will not merge two nearby neurons with relatively low inadaptability. This is because the lower inadaptability indicates a clear distribution boundary between the two microclusters.
\end{remark}

Through the SGM training, a network $A$ containing $Q$ neurons organized in a certain number of subnetworks is obtained. Each neuron can be treated as a microcluster center to partition the data chunk into $Q$ microclusters. $A$ is also utilized for quickly retrieving the neighboring neurons, which is significant in accelerating the hierarchical microcluster merging process in the next subsection.

\subsection{HM: Hierarchical Merging for Imbalanced Clustering}

With the distribution knowledge of the current data chunk obtained through SGM training, the microclusters are merged to obtain a proper number of imbalanced clusters. The topological structure of the neurons is exploited to accelerate the merging process. Since clusters corresponding to non-adjacent neurons are less likely to be merged, 
the topology is utilized as a retrieval structure that only allows the merging of adjacent neurons, thereby significantly avoiding the laborious traversing of all the possible microcluster pairs during the merging.

To judge the merging of two clusters, the concept of density gap is introduced to describe the prominence of their boundary. Based on this, global compactness and global separability can be derived as measures to monitor the merging process and guide the selection of the optimal number of clusters. Specifically, $Q$ microclusters $\{G_1, G_2, ..., G_Q\}$ obtained through SGM are the initial microclusters. During their merging, the current $k$ clusters composed of a certain number of microclusters are denoted as $\varPhi(k) = \{C_1, C_2, ..., C_k\}$, with their centers denoted as $S = \{\mathbf{s}_1, \mathbf{s}_2, \dots, \mathbf{s}_k\}$. Note that these cluster centers are not necessarily original neurons, as each cluster may consist of data objects belonging to multiple neurons' corresponding microclusters. Given two clusters $C_i$ and $C_j$, their merging is considered by projecting all their objects onto the 1-D space passing through their centers $\mathbf{s}_{i}$ and $\mathbf{s}_{j}$:
\begin{equation}
    \mathbf{x}' = \frac{(\mathbf{x} - \bar{\mathbf{s}})^T(\mathbf{s}_{i} - \mathbf{s}_{j})}{\lVert \mathbf{s}_{i} - \mathbf{s}_{j} \rVert_2},
    \label{13}
\end{equation}
where $\bar{\mathbf{s}}=\left( \mathbf{s}_{i} +\mathbf{s}_{j} \right) /2$. To characterize the 1-D Gaussian mixture probability density distribution of $C_i$ and $C_j$, their objects are further mapped onto the 1-D space with respective cluster centers as 0.5 and -0.5:
\begin{equation}
 \resizebox{0.98\columnwidth}{!}{$
    M\left( u;i,j \right) = \dfrac{\text{card}(C_i) f( u|0.5,\sigma _{i}^{2}) +\text{card}(C_j)f( u|-0.5,\sigma _{j}^{2}) }{\text{card}(C_i)+\text{card}(C_j)}
 $},
\end{equation}
where $f(u|0.5,\sigma_i^2)$ describes the probability density distribution of $C_i$, with variance $\sigma_i^2$ computed from the objects in $C_i$ after mapping them to the 1-D space crossing $\mathbf{s}_{i}$ and $\mathbf{s}_{j}$. Then the density gap between $C_i$ and $C_j$ can be defined as:
\begin{equation}
    m_{ij}=\frac{1}{\underset{u\in U}{\min}M\left( u;i,j \right)},
    \label{15}
\end{equation}
where $U=\{-0.5, -0.49, \ldots, 0.5\}$ is a traverse set of $u$ with a step size of 0.01. Since a smaller $\min_{u\in U} M(u;i,j)$ reflects a more prominent distribution density gap between two clusters, then the corresponding $m_{ij}$ will be larger, indicating that $C_i$ and $C_j$ are more unsuitable for merging. Conversely, $C_i$ and $C_j$ are selected for merging if their $m_{ij}$ reaches the global minimum among all the cluster pairs in $\varPhi(k)$, which can be defined as the global compactness:
\begin{equation}
\theta_{k} = \underset{\substack{C_i, C_j \in \varPhi(k)}}{\min} m_{ij}.
\label{16}
\end{equation}
It reflects the compactness of the clusters that will be merged currently in $\varPhi(k)$. A lower $\theta_{k}$ indicates that the new cluster formed by merging will be more compact, and thus a lower $\theta_{k}$ is preferred by the typical clustering objective. After the merging, a new cluster is formed as $C_v=\{C_i,C_j\}$ and added to $\varPhi(k)$, while the original $C_i$ and $C_j$ are removed:
\begin{equation}
\varPhi(k)\setminus\{C_i,C_j\}\ \ \text{and}\ \ \varPhi(k)\cup C_v.
\label{17}
\end{equation}
Then a new status $\varPhi(k-1)$ arises as the number of clusters becomes $k-1$. The merging stops when all clusters are merged into one cluster, i.e., $k = 1$. 

An optimal $k$ can be selected by evaluating the merging process. Since $\theta_{k}$ may keep monotonic increasing due to the gradual merging of adjacent microclusters with lower density gaps, it is incompetent in determining the optimal number of clusters. Therefore, a global separability is also introduced:
\begin{equation}
    \omega_k=\underset{C_i\in \varPhi(k)}{\max}\left(\sum_{\mathbf{x}_j\in C_i}{\frac{N_s(\mathbf{x}_j)\setminus\{N_s(\mathbf{x}_j)\cap C_i\}}{s}}\right),
    \label{18}
\end{equation}
where $s$ refers to the number of global nearest neighbors of an object $\mathbf{x}_j\in C_i$, and $N_s(\mathbf{x}_j)\setminus\{N_s(\mathbf{x}_j)\cap C_i\}$ represents the number of objects that are the $s$-nearest neighbors of $\mathbf{x}_j$ but are not included in the cluster $C_i$. It is intuitive that if a cluster $C_i$ is located far away from the other clusters, a small $(N_s(\mathbf{x}_j)\setminus\{N_s(\mathbf{x}_j)\cap C_i\})/s$ is yielded as most $s$-nearest neighbors of $\mathbf{x}_j$ may be included in its cluster $C_i$. The cluster size of $C_i$ is not introduced to this measure to ensure equal status to imbalanced clusters, thus avoiding overlooking relatively small clusters in ISDC. Accordingly, a lower $\omega_k$ indicates a higher separability of the current clusters, which is preferred by the typical clustering objective. As merging progresses, $\omega_k$ decreases because more inseparable adjacent clusters are merged to form larger prominent clusters.

By collectively considering the global compactness and global separability, a proper number of clusters can be automatically determined as:
\begin{equation}
\resizebox{1\columnwidth}{!}{$
\hat{k}^* = \underset{k\in\{1,2,...,Q\}}{\arg\min}\left( \dfrac{\theta_k}{\max(\theta_1,\theta_2,...,\theta_{Q})} + \dfrac{\omega_k}{\max(\omega_1,\omega_2,...,\omega_{Q})} \right),$}
\label{19}
\end{equation}
where the two denominators are utilized to normalize $\theta_k$ and $\omega_k$ into the same scale to ensure their comparability, and form a trade-off between them. Consequently, the optimal number of clusters $k^*$ is estimated as the knee point \cite{lu2019self}, i.e., $\hat{k}^*$, of the composite curve of $\theta_k$ and $\omega_k$.

\subsection{Overall Algorithm and Complexity Analysis}

\begin{algorithm}[!t]
\caption{SOHI: Self-grOwth maps-guided Hierarchical merging for Imbalanced data clustering}\label{A1}
\begin{algorithmic}
\raggedright %
\setlength{\rightskip}{1cm} 
\Statex \hspace{-0.4cm} \textbf{Input:} Data chunk $X^\iota$, learning rates $\epsilon_{\Omega}$, $\epsilon_b$, reduction coefficient $\alpha$, and number of nearest neighbors $s$.
\Statex \hspace{-0.4cm} \textbf{Output:} $\hat{k}^*$ clusters $\{C_1, C_2, ..., C_{\hat{k}^*}\}$.
\begin{enumerate}[label=\small{\arabic*.}, leftmargin=0cm]
\item \textbf{SGM: Self-Growth Maps:}
\begin{enumerate}[label=\small{\theenumi\arabic*}]
\setlength{\rightskip}{0.3cm} 
\item Initialize network $A$ with $T$ subnetworks;
\item Train $A$ according to Eqs. (\ref{n7})-(\ref{n7_1}) by randomly selecting objects from $X^\iota$;
\item Update $A$ with created neurons or merged subnetworks according to Algorithm ~\ref{Alg_IGCM}.
\end{enumerate}
\setlength{\rightskip}{0.3cm} 
\item \textbf{HM: Hierarchical Merging:}
\begin{enumerate}[label=\small{\theenumi\arabic*}]
\setlength{\rightskip}{0.3cm} 
\item Calculate density gap of adjacent microclusters according to Eqs. (\ref{13})-(\ref{15});
\item Compute global compactness according to Eq. (\ref{16}) and merge all the clusters through $A$-guided retrieval;
\item Compute global separability according to Eq. (\ref{18});
\item Obtain $\hat{k}^*$ clusters according to Eq. (\ref{19}). 
\end{enumerate}
\end{enumerate}
\end{algorithmic}
\end{algorithm}

The complete SOHI is summarized as Algorithm \ref{A1}. SGM is first implemented to grow and merge subnetworks to represent the distribution of the imbalanced dataset. Then, the microclusters corresponding to the neurons are adaptively merged using HM, and an optimal number of clusters is obtained by evaluating the merging process. The time and space complexity of SOHI is analyzed in the following.

\begin{theorem}\label{the:complexity_adc}
\textbf{Time Complexity:} Given an $n$-object chunk $X^\iota$, and its $Q$-neuron SGM. The time complexity of SOHI is $O(nQ^2d)$.
\end{theorem}
\begin{prf}
Time complexity of SGM: The growth of SGM is driven by all the $n$ objects. For each object, the time complexity is analyzed as follows: The distances between the object and all $Q$ neurons are computed to find the BMN, which incurs a complexity of $O(Qd)$, where $d$ denotes the feature dimensionality. The BMN and its $B$ adjacent neurons are updated, costing $O(Bd)$, where $B$ is the branching factor of the network. The inadaptability of the BMN and at most $Q$ neurons in the same subnetwork are also updated, resulting in a complexity of $O(d+Q)$. Therefore, updating the neurons and their inadaptabilities for all $n$ objects results in a time complexity of $O(nBd+ nQ+ nd)$.
Since the neuron creation or subnetworks merging of $A$ is triggered after every $\rho$ objects are input, Algorithm~\ref{Alg_IGCM} will be implemented $\frac{n}{\rho}$ times in total. The time complexity for each implementation is analyzed as follows: Identifying the neurons with the largest and second-largest inadaptabilities, e.g., $\mathbf{z}_{i,v}$ and $\mathbf{z}_{j,g}$, requires $O(Q)$ time (line 1 of Algorithm \ref{Alg_IGCM}), as the inadaptability of each neuron has been prepared. Then, finding the furthest adjacent neuron $\mathbf{z}_{i,q}$ among the $B$ neighbors of $\mathbf{z}_{i,v}$ takes $O(Bd)$ time (line 2). Computing the distances for judging the condition (line 3) takes $O(2d)$ time. Therefore, $O(Q + Bd + 2d)$ is required to determine whether a new neuron should be created or subnetworks should be merged. If a new neuron $\mathbf{z}_{i,h}$ is to be created (lines 3–6), it takes $O(2d + Bd)$ to initialize the neuron as the mean of the two adjacent neurons and to update the inadaptability according to its $B$ neighbors. The $B$ neighbors of the new neurons should also be updated, each of which involves $O(Bd)$ operations. Thus, the total cost for $B$ neighbor updating is $O(B^2d)$. Hence, the complexity for a new neuron creation is $O(2d + Bd + B^2d)$. If subnetwork merging is triggered (lines 8) to connect three neurons, a negligible computational complexity is involved. Consequently, the worst-case time complexity for one iteration of Algorithm~\ref{Alg_IGCM} is $O(Q + Bd + 2d + 2d + Bd + B^2d)$, and for all $\frac{n}{\rho}$ implementations, it will be $O(\frac{n}{\rho}Q + \frac{n}{\rho}B^2d)$.
In summary, the overall time complexity of the SGM process is $O(nQ+ nd+ nB^2d)$.

Time complexity of HM: The SGM topology is leveraged to accelerate the merging, and we analyze the complexity in two main stages: 1) IntrA-Subnetwork Merging (IASM), and 2) IntEr-Subnetwork Merging (IESM). 
In IASM, at most $Q$ microclusters should be considered within the same subnetwork for merging. Each merge estimates a density distribution based on at most $n$ objects from two candidate clusters with complexity $O(nd)$. The two candidates are chosen based on the density gap between all possible pairs of clusters. Since there are at most $Q$ microclusters, and for each microcluster, only the adjacent $B$ microclusters are considered for merging, the complexity is thus $O(nQBd)$. Searching for the cluster pair with the smallest gap involves complexity $O(QB)$. Since there are at most $Q$ merges, the time complexity is $O(Q(nd+nQBd+QB))=O(nQd+nQ^2Bd+Q^2B)$, which can be simplified to $O(nQ^2Bd)$. 
For IESM, merging the current clusters corresponding to the subnetworks should be conducted at most $T$ times. Each cluster contains at most $n$ objects, and each merge requires up to $T^2$ density gap computations with complexity $O(nT^2d)$. The overall IESM complexity is thus $O(nT^3d)$. In addition to the above two stages, each merge also involves calculating global separability. This process is analyzed as follows: The most similar $s$ neighbors of each object should be identified by computing pairwise similarities of $n$ objects, which incurs a complexity of $O(n^2d)$. Then the global separability can be computed by traversing all $n$ objects and their $s$ neighbors, which takes $O(ns)$ complexity. Therefore, the overall complexity for global separability computation is $O(n^2d + ns)$. Since at most $Q$ merges are conducted to merge all $Q$ microclusters, the overall complexity for the separability computation is $O(Q(n^2d + ns))=O(n^2Qd+nQs)$. An alternative efficient way for computing the separability is to treat neurons as objects. Specifically, it requires finding the $s$ neighbors of each of the $Q$ neurons based on the $Q\times Q$ similarity matrix, incurring a time complexity of $O(Q^3d+Q^2s)$ for $Q$ merges in total. In summary, the total time complexity of the HM process is $O(nQ^2Bd+ nT^3d+ Q^3d+ Q^2s)$, which can be simplified to $O(nQ^2Bd)$, given that $Q\ll n$, and both $T$ and $s$ are small constants in most cases. 

The overall time complexity of SOHI, combining the complexity of SGM and HM, is $O(nQ+ nd + nB^2d + nQ^2Bd)$. Since $B$ is also a small constant, the time complexity of SOHI can be simplified to $O(nQ^2d)$.

\qed
\end{prf}

\begin{theorem}\label{the:complexity_space}
\textbf{Space Complexity:} Given an $n$-object chunk $X^\iota$, and its $Q$-neuron SGM. The space complexity of SOHI is $S(nd + nQ)$.
\end{theorem}
\begin{prf}
Space complexity of SGM: The growth of SGM involves the storage of $X^\iota$ as an $n\times d$ matrix. The SGM is with $Q$ neurons described by $d$-dimensional vectors, and each neuron is connected to at most $B$ neighbors, where $B$ is the branching factor. Thus, the SGM can be described by a $Q\times d$ matrix and a $Q\times B$ adjacency matrix for neuron and linkage description, respectively. The object-neuron similarity is stored in an $n\times Q$ matrix. A $Q$-dimensional vector is also required to record the inadaptability of the $Q$ neurons. Therefore, the overall space complexity of SGM is $S(nd + Qd + QB + nQ + Q)$.

Space complexity of HM: Based on $X^\iota$, SGM, and the object-neuron similarity matrix stored in the SGM phase, HM needs additional space to store the density gaps between at most $Q(Q-1)/2$ pairs of clusters during merging. Global separability computation at the object level needs the storage of $s$ neighbors for each of the $n$ objects. Accordingly, the space complexity required by HM is $S(Q^2+ns)$. 

The overall space complexity of SOHI is thus $S(nd+Qd+QB+nQ+Q+Q^2+ns)$, which can be simplified to $S(nd+nQ)$, given that $Q \ll n$, and $B$ and $s$ are small constants in most cases.
\qed 
\end{prf}

\begin{algorithm}[!t]
\caption{TLRS: Two-Layer Random Sampling}\label{ATLRS}
\SetAlgoLined
\KwIn{The whole original dataset $X$, the maximum imbalance ratio $IR$, true number of clusters $k^*$, true clustering partition $C$.}
\KwOut{Data chunk $X^\iota$ and its clustering partition ${C}^\iota$.}
Randomly set cluster number as an integer : $2 \leq k^\iota \leq k^*$; Initialize a imbalance ratio set $IR^\iota$ as empty arrays\;

\For{$i \gets 1$ to $k^\iota-1$}{
     $IR^\iota_{i} \gets \text{DiscreteUniform}([1, IR])$\;
}
$IR^\iota \gets$ sort $IR^\iota$ in ascending order\;
$\text{card}(C^\iota_1)\gets \text{card}(C_1)$\;
\For{$i \gets 2$ to $k^\iota$}{
    \uIf{$IR_{i-1}^{\iota} \cdot \text{card}(C_{i-1}) \leq \text{card}(C_{i})$}{
        $\text{card}(C^\iota_i) \gets IR^\iota_{i-1} \cdot \text{card}(C_{i-1})$\;
        
    }\Else{
        $\text{card}(C^\iota_i) \gets \text{card}(C_{i})$\;
    }
}
$C^\iota_i$ data objects are randomly taken from the $i$-th cluster and form $X^\iota$.
\end{algorithm}

\section{Experiments}\label{section5}

Three experiments, i.e., efficiency evaluation, clustering accuracy evaluation, and ablation study, have been conducted on eleven datasets by comparing ten counterparts, including eight existing methods and two ablated versions of SOHI. 

\begin{table}[!t]
\caption{Statistical information of the eleven datasets. $d$, $n$, $k^*$, and $IR$ represent the numbers of attributes, data objects, true clusters, and the imbalance ratio, respectively.}
\label{tab:datasets}
\centering
\begin{tabular}{c|cc|rrrr} 
\toprule
No. & Dataset   & Abbrev.      & $d$ & $n$  & $IR$    &  $k^*$ \\
\midrule
1 &Gaussian     &GA          & 2          & 2000      & 19.87        & 4          \\
2 &IDS2         &ID        & 2          & 3200      & 10.00        & 5          \\
3 &Abalone      &AB          & 8          & 4177      & 689.00       & 28          \\
4 &Car Evaluation&CE          & 6          & 1728      & 18.62        & 4         \\
5 &Haberman's Survival     &HS           & 3          & 306       & 2.78         & 2          \\
6 &Heart Failure     &HF         & 12         & 299       & 2.11         & 2          \\
7 &Land Mines   &LM           & 3          & 338       & 1.09         & 5          \\
8 &Page Blocks  &PB           & 10         & 5473      & 175.43       & 5          \\
9 &Raisin       &RA           & 7          & 900       & 1.00         & 2          \\
10 &Seeds       &SE          & 7          & 210       & 1.00         & 3          \\
11 &Wholesale Customers &WC          & 7          & 440       & 6.72         & 3          \\
\bottomrule
\end{tabular}
\end{table}

\begin{table*}[!t]
\caption{Clustering performance of different methods evaluated by the ARI ($\uparrow$), NMI ($\uparrow$) and DBI ($\downarrow$) metrics. The results marked in \textbf{\textcolor{orange}{Orange}} and \textbf{\textcolor{gray}{Gray}} colors indicate the best and second-best results on each dataset, respectively.}
\label{tb:result}
\centering
\resizebox{\textwidth}{!}{
\begin{tabular}{c|c|cccccccc|c}
\toprule
\textbf{Dataset} & \textbf{Metric} &\textbf{BIRCH}\ \ \ \ \ 	&\textbf{StreamKM++}	&\textbf{CPCL}\ \ \ \ \ 	&\textbf{SMCL}\ \ \ \ \ &\textbf{IGMTT}\ \ \ \  &\textbf{DenSOINN} \    &\textbf{LDPI} \ \ \ \   &\textbf{M3W}\ \ \ \ \  &\textbf{SOHI} (ours)\\
\hline
\multirow{3}{*}{GA} & ARI & 0.2360$\pm$0.00 & 0.0098$\pm$0.01 & -0.0002$\pm$0.01\phantom{-} & \cellcolor{gray!20}0.9252$\pm$0.06 & 0.5725$\pm$0.22 & 0.3234$\pm$0.01 & 0.0226$\pm$0.07 & 0.8264$\pm$0.13 & \cellcolor{orange!40}0.9667$\pm$0.01\\
    & NMI & 0.0053$\pm$0.00 & 0.0122$\pm$0.00 & 0.0114$\pm$0.00 & \cellcolor{gray!20}0.8685$\pm$0.08 & 0.6744$\pm$0.12 & 0.4707$\pm$0.00 & 0.0480$\pm$0.12 & 0.8406$\pm$0.07 & \cellcolor{orange!40}0.9447$\pm$0.02\\
    & DBI & 25.0641$\pm$2.85\phantom{1} & 0.7303$\pm$0.14 & 0.9589$\pm$0.16 & \cellcolor{gray!20}0.5321$\pm$0.05 & 0.9871$\pm$0.35 & 1.3108$\pm$4.82 & 0.9864$\pm$0.28 & 3.6246$\pm$1.91 & \cellcolor{orange!40}0.4879$\pm$0.01\\
\midrule

\multirow{3}{*}{ID} & ARI & 0.0008$\pm$0.00 & 0.0077$\pm$0.00 & 0.0118$\pm$0.01 & \cellcolor{orange!40}1.0000$\pm$0.00 & 0.6611$\pm$0.10 & 0.0055$\pm$0.00 & 0.5959$\pm$0.00 & 0.8528$\pm$0.00 & \cellcolor{gray!20}0.9930$\pm$0.01\\
    & NMI & 0.0043$\pm$0.00 & 0.0085$\pm$0.00 & 0.0168$\pm$0.00 & \cellcolor{orange!40}1.0000$\pm$0.00 & 0.7668$\pm$0.06 & 0.0097$\pm$0.00 & 0.6983$\pm$0.00 & 0.8367$\pm$0.00 & \cellcolor{gray!20}0.9901$\pm$0.01\\
    & DBI & 52.6033$\pm$13.26 & 0.2214$\pm$0.00 & 0.8440$\pm$0.05 & \cellcolor{orange!40}0.2084$\pm$0.00 & 0.7807$\pm$0.16 & 50.8885$\pm$8.07\phantom{-} & 0.2198$\pm$0.00 & 1.4958$\pm$0.01 & \cellcolor{gray!20}0.2194$\pm$0.01\\
\midrule

\multirow{3}{*}{AB} & ARI & 0.0027$\pm$0.00 & 0.0610$\pm$0.17 & 0.0031$\pm$0.01 & -0.2316$\pm$0.01\phantom{-} & 0.0928$\pm$0.04 & 0.0001$\pm$0.00 & 0.0010$\pm$0.01 & \cellcolor{orange!40}0.2153$\pm$0.03 & \cellcolor{gray!20}0.1431$\pm$0.03 \\
         & NMI & 0.0862$\pm$0.01 & 0.1741$\pm$0.21 & 0.0795$\pm$0.03 & 0.2140$\pm$0.07 &0.2535$\pm$0.05 & 0.0650$\pm$0.01 & 0.0289$\pm$0.00 & \cellcolor{orange!40}0.4147$\pm$0.05 & \cellcolor{gray!20}0.2819$\pm$0.07 \\
         & DBI & 0.8233$\pm$0.04 & 0.7427$\pm$0.08 & 0.8835$\pm$0.45 & 0.9685$\pm$0.16 & 1.6856$\pm$0.18 & 4.4526$\pm$1.97 & \cellcolor{orange!40}0.6181$\pm$0.13 & 2.2018$\pm$1.66 & \cellcolor{gray!20}0.6918$\pm$0.26\\
\midrule

\multirow{3}{*}{CE}  & ARI & -0.0004$\pm$0.00\phantom{-} & \cellcolor{gray!20}0.0847$\pm$0.00 & -0.0076$\pm$0.02\phantom{-} & -0.0895$\pm$0.37\phantom{-} & 0.0069$\pm$0.02 & -0.0010$\pm$0.00\phantom{-} & 0.0092$\pm$0.02 & \multicolumn{1}{c|}{$-$} & \cellcolor{orange!40}0.1138$\pm$0.07\\
    & NMI & 0.0052$\pm$0.00 & 0.0139$\pm$0.00 & 0.0120$\pm$0.01 & 0.0424$\pm$0.24 & \cellcolor{gray!20}0.0652$\pm$0.01 & 0.0052$\pm$0.00 & 0.0243$\pm$0.01 & \multicolumn{1}{c|}{$-$} & \cellcolor{orange!40}0.0837$\pm$0.06\\
    & DBI & 20.8972$\pm$1.61\phantom{1} & \cellcolor{orange!40}1.4144$\pm$0.01 & 3.0599$\pm$0.71 & 3.5717$\pm$1.43 & \cellcolor{gray!20}1.7767$\pm$0.10 & 20.8279$\pm$1.58\phantom{1} & 1.8903$\pm$0.21 & \multicolumn{1}{c|}{$-$} & 1.9068$\pm$0.38\\
\midrule

\multirow{3}{*}{HS} & ARI & -0.0028$\pm$0.01\phantom{-} & \cellcolor{gray!20}0.0465$\pm$0.02 & 0.0059$\pm$0.03 & -0.0166$\pm$0.01\phantom{-} & 0.0018$\pm$0.01 & 0.0066$\pm$0.01 & -0.0022$\pm$0.03\phantom{-} & -0.0101$\pm$0.02\phantom{-} & \cellcolor{orange!40}0.0613$\pm$0.03 \\
       & NMI & 0.0046$\pm$0.00 & 0.0198$\pm$0.01 & 0.0140$\pm$0.01 & 0.0077$\pm$0.00 & 0.0071$\pm$0.00 & 0.0105$\pm$0.00 & \cellcolor{gray!20}0.0202$\pm$0.02 & 0.0127$\pm$0.01 & \cellcolor{orange!40}0.0251$\pm$0.01\\
      & DBI & 9.5495$\pm$1.91 & 0.9385$\pm$0.01 & 1.3509$\pm$0.26 & 0.8764$\pm$0.05 & 1.5378$\pm$0.56 & 15.8149$\pm$3.38\phantom{1} & \cellcolor{orange!40}0.3927$\pm$0.15 & 0.8223$\pm$0.01 & \cellcolor{gray!20}0.7448$\pm$0.16\\
\midrule

\multirow{3}{*}{HF} & ARI & 0.0039$\pm$0.01 & \cellcolor{gray!20}0.0333$\pm$0.02 & 0.0200$\pm$0.03 & 0.0058$\pm$0.01 & 0.0046$\pm$0.02 & 0.0041$\pm$0.00 & \multicolumn{1}{c}{$-$} & 0.0035$\pm$0.00 & \cellcolor{orange!40}0.0346$\pm$0.03\\
     & NMI & 0.0049$\pm$0.00 & \cellcolor{gray!20}0.0268$\pm$0.02 & 0.0106$\pm$0.01 & 0.0038$\pm$0.00 & 0.0047$\pm$0.00 & 0.0080$\pm$0.00 & \multicolumn{1}{c}{$-$} & \cellcolor{orange!40}0.0389$\pm$0.00 & 0.0117$\pm$0.01\\
     & DBI & 27.3766$\pm$19.03 & \cellcolor{orange!40}0.3158$\pm$0.26 & 0.6165$\pm$0.03 & 1.9381$\pm$0.80 & 2.2353$\pm$0.14 & 46.5634$\pm$19.15 & \multicolumn{1}{c}{$-$} & 1.7595$\pm$0.00 & \cellcolor{gray!20}0.5804$\pm$0.11\\
\midrule

\multirow{3}{*}{LM} & ARI & 0.0010$\pm$0.00 & 0.0130$\pm$0.01 & \cellcolor{gray!20}0.0585$\pm$0.01 & 0.0419$\pm$0.02 & 0.0147$\pm$0.02 & 0.0028$\pm$0.00 & 0.0059$\pm$0.01 & 0.0189$\pm$0.03 & \cellcolor{orange!40}0.1066$\pm$0.07\\
    & NMI & 0.0179$\pm$0.00 & 0.0311$\pm$0.01 & 0.1037$\pm$0.01 & \cellcolor{gray!20}0.1660$\pm$0.04 & 0.0446$\pm$0.04 & 0.0182$\pm$0.00 & 0.0405$\pm$0.02 & 0.1262$\pm$0.06 & \cellcolor{orange!40}0.2033$\pm$0.05\\
    & DBI & 17.8274$\pm$2.17\phantom{1} & 0.9159$\pm$0.03 & \cellcolor{gray!20}0.8714$\pm$0.02 & 0.9369$\pm$0.04 & 0.9819$\pm$0.04 & 18.4434$\pm$2.68\phantom{1} & \cellcolor{orange!40}0.8165$\pm$0.12 & 1.8742$\pm$0.34 & 0.9554$\pm$0.31\\
\midrule

\multirow{3}{*}{PB} & ARI & -0.0011$\pm$0.00\phantom{-} & 0.0112$\pm$0.01 & -0.0059$\pm$0.01\phantom{-} & -0.0323$\pm$0.01\phantom{-} & \cellcolor{gray!20}0.1224$\pm$0.04 & 0.0034$\pm$0.00 & 0.0103$\pm$0.01 & 0.0315$\pm$0.03 & \cellcolor{orange!40}0.1336$\pm$0.08\\
    & NMI & 0.0123$\pm$0.00 & 0.0282$\pm$0.01 & 0.0076$\pm$0.00 & 0.0382$\pm$0.01 & \cellcolor{gray!20}0.1936$\pm$0.04 & 0.0170$\pm$0.00 & 0.0131$\pm$0.01 & \cellcolor{orange!40}0.2376$\pm$0.01 & 0.1491$\pm$0.04\\
    & DBI & 5.5314$\pm$0.74 & 0.9143$\pm$0.02 & 0.6875$\pm$0.09 & 1.1289$\pm$0.04 & 1.0646$\pm$0.20 & 17.8086$\pm$1.16\phantom{1} & \cellcolor{orange!40}0.4487$\pm$0.11 & 1.174$\pm$0.10 &  \cellcolor{gray!20}0.5160$\pm$0.27\\
\midrule

\multirow{3}{*}{RA} & ARI & 0.0009$\pm$0.01 & 0.0361$\pm$0.02 & 0.0070$\pm$0.01 & \cellcolor{gray!20}0.5173$\pm$0.22 & 0.2135$\pm$0.10 & 0.0137$\pm$0.01 & 0.0053$\pm$0.01 & 0.0642$\pm$0.00 & \cellcolor{orange!40}0.5319$\pm$0.02 \\
        & NMI & 0.0027$\pm$0.00 & 0.0083$\pm$0.01 & 0.0124$\pm$0.00 & \cellcolor{gray!20}0.4568$\pm$0.22 & 0.2024$\pm$0.08 & 0.0105$\pm$0.00 & 0.0020$\pm$0.00 & 0.1757$\pm$0.00 & \cellcolor{orange!40}0.3979$\pm$0.01\\
        & DBI & 21.2803$\pm$9.64\phantom{1} & 0.5071$\pm$0.00 & 0.5644$\pm$0.01 & 0.9011$\pm$0.14 & 1.6233$\pm$0.57 & 84.9281$\pm$22.33 & \cellcolor{orange!40}0.3169$\pm$0.00 & 5.3499$\pm$0.00 & \cellcolor{gray!20}0.3607$\pm$0.11\\
\midrule

\multirow{3}{*}{SE} & ARI & 0.0023$\pm$0.01 & 0.0648$\pm$0.02 & 0.0848$\pm$0.01 & \cellcolor{gray!20}0.6047$\pm$0.14 & 0.3277$\pm$0.13 & 0.0039$\pm$0.01 & 0.0103$\pm$0.06 & 0.3040$\pm$0.12 & \cellcolor{orange!40}0.8684$\pm$0.05\\
    & NMI & 0.0260$\pm$0.00 & 0.0600$\pm$0.01 & 0.1218$\pm$0.00 & \cellcolor{gray!20}0.6790$\pm$0.09 & 0.4457$\pm$0.11 & 0.0236$\pm$0.00 & 0.0344$\pm$0.04 & 0.4591$\pm$0.05 & \cellcolor{orange!40}0.7920$\pm$0.07\\
    & DBI & 13.4789$\pm$2.73\phantom{1} & 0.8175$\pm$0.06 & 0.9336$\pm$2.73 & 0.8401$\pm$0.11 & 1.6062$\pm$0.45 & 12.0563$\pm$2.43\phantom{1} & \cellcolor{orange!40}0.6264$\pm$0.03 & 1.2176$\pm$0.13 & \cellcolor{gray!20}0.6531$\pm$0.09\\
\midrule

\multirow{3}{*}{WC} & ARI & 0.0004$\pm$0.00 & 0.0205$\pm$0.01 & -0.0017$\pm$0.01\phantom{-} & \cellcolor{orange!40}0.0535$\pm$0.02 & \cellcolor{gray!20}0.0420$\pm$0.01 & 0.0032$\pm$0.00 & -0.0025$\pm$0.01\phantom{-} & -0.0033$\pm$0.00\phantom{-} & 0.0036$\pm$0.00\\
    & NMI & 0.0088$\pm$0.00 & 0.0283$\pm$0.01 & 0.0141$\pm$0.01 & \cellcolor{orange!40}0.0501$\pm$0.01 & 0.0418$\pm$0.02 & 0.0124$\pm$0.00 & 0.0125$\pm$0.00 & 0.0179$\pm$0.00 & \cellcolor{gray!20}0.0473$\pm$0.01\\
    & DBI & 12.1115$\pm$1.56\phantom{1} & 0.8360$\pm$0.08 & 1.1122$\pm$0.11 & \cellcolor{gray!20}0.5299$\pm$0.06 & 1.0990$\pm$0.12 & 11.1380$\pm$1.84\phantom{1} & 1.2536$\pm$0.08 & 0.6666$\pm$0.07 & \cellcolor{orange!40}0.3799$\pm$0.11\\
\bottomrule
\end{tabular}}
\end{table*}

\subsection{Experimental Settings}
Eleven datasets, including two synthetic and nine real datasets with varying sizes, dimensions, and distribution types, are utilized for the experiments. Their statistics are provided in Table~\ref{tab:datasets}. ID and GA \cite{liang} are synthesized by applying a mixture of bivariate Gaussian density functions~\cite{lu2019self}. All real datasets are obtained from the UCI Machine Learning Repository \cite{kelly2021uci}. Min-max normalization is adopted to pre-process each feature into the identical value domain $[0,1]$.

\begin{table}[!t]
\centering
\caption{Comparison of ablated SOHI variants. The symbol ``$✓$'' indicates that the corresponding component is not ablated. The symbol ``\textcolor{red}{\ddag}'' represents the expected performance degradation compared to SOHI (i.e., the first row on each dataset).}
\label{tb:ablation}
\setlength{\tabcolsep}{2pt}
    \begin{tabular}{c|ccc|r|r|r}
\toprule
Dataset                     & MS & SGM & HM & \multicolumn{1}{c|}{ARI} & \multicolumn{1}{c|}{NMI} & \multicolumn{1}{c}{DBI} \\  
\midrule
\multirow{3}{*}{GA}   & ✓ & ✓ & ✓ & 0.9674$\pm$0.01\textcolor{white}{\ddag} & 0.9228$\pm$0.02\textcolor{white}{\ddag} & 0.4969$\pm$0.02\textcolor{white}{\ddag} \\
                        & & ✓ & ✓ & 0.8207$\pm$0.25\textcolor{red}{\ddag} & 0.7996$\pm$0.20\textcolor{red}{\ddag} & 0.5174$\pm$0.08\textcolor{red}{\ddag} \\
                          & & & ✓ & 0.7298$\pm$0.07\textcolor{red}{\ddag} & 0.6746$\pm$0.05\textcolor{red}{\ddag} & 1.0782$\pm$0.18\textcolor{red}{\ddag} \\ \midrule

\multirow{3}{*}{ID}   & ✓ & ✓ & ✓ & 0.9958$\pm$0.09\textcolor{white}{\ddag} & 0.9889$\pm$0.05\textcolor{white}{\ddag} & 0.4154$\pm$0.07\textcolor{white}{\ddag} \\
                        & & ✓ & ✓ & 0.8357$\pm$0.03\textcolor{red}{\ddag} & 0.8832$\pm$0.02\textcolor{red}{\ddag} & 0.5835$\pm$0.08\textcolor{red}{\ddag} \\
                          & & & ✓ & 0.6516$\pm$0.00\textcolor{red}{\ddag} & 0.7029$\pm$0.00\textcolor{red}{\ddag} & 0.7464$\pm$0.00\textcolor{red}{\ddag} \\ \midrule

\multirow{3}{*}{AB}   & ✓ & ✓ & ✓ & 0.0386$\pm$0.00\textcolor{white}{\ddag} & 0.0910$\pm$0.00\textcolor{white}{\ddag} & 0.5494$\pm$0.11\textcolor{white}{\ddag} \\
                        & & ✓ & ✓ & 0.0385$\pm$0.00\textcolor{red}{\ddag} & 0.0907$\pm$0.00\textcolor{red}{\ddag} & 1.1043$\pm$0.04\textcolor{red}{\ddag}  \\
                          & & & ✓ & 0.0048$\pm$0.00\textcolor{red}{\ddag} & 0.0347$\pm$0.00\textcolor{red}{\ddag} & 0.6926$\pm$0.00\textcolor{red}{\ddag} \\ \midrule
                            
\multirow{3}{*}{CE}   & ✓ & ✓ & ✓ & 0.1087$\pm$0.03\textcolor{white}{\ddag} & 0.0832$\pm$0.02\textcolor{white}{\ddag} & 1.6209$\pm$0.02\textcolor{white}{\ddag} \\
                        & & ✓ & ✓ & 0.0258$\pm$0.08\textcolor{red}{\ddag} & 0.0223$\pm$0.04\textcolor{red}{\ddag} & 1.8348$\pm$0.10\textcolor{red}{\ddag} \\
                          & & & ✓ & 0.0151$\pm$0.00\textcolor{red}{\ddag} & 0.0072$\pm$0.00\textcolor{red}{\ddag} & 2.4099$\pm$0.00\textcolor{red}{\ddag}\\ \midrule

\multirow{3}{*}{HS}   & ✓ & ✓ & ✓ & 0.1357$\pm$0.04\textcolor{white}{\ddag} & 0.0647$\pm$0.02\textcolor{white}{\ddag} & 0.3658$\pm$0.03\textcolor{white}{\ddag} \\
                        & & ✓ & ✓ & 0.1103$\pm$0.01\textcolor{red}{\ddag} & 0.0468$\pm$0.00\textcolor{red}{\ddag} & 0.6296$\pm$0.18\textcolor{red}{\ddag} \\
                          & & & ✓ & 0.0143$\pm$0.00\textcolor{red}{\ddag} & 0.0051$\pm$0.00\textcolor{red}{\ddag} & 1.6586$\pm$0.02\textcolor{red}{\ddag} \\ \midrule
                            
\multirow{3}{*}{HF}   & ✓ & ✓ & ✓ & 0.0475$\pm$0.00\textcolor{white}{\ddag} & 0.0206$\pm$0.01\textcolor{white}{\ddag} & 0.1242$\pm$0.01\textcolor{white}{\ddag}\\
                        & & ✓ & ✓ & 0.0053$\pm$0.00\textcolor{red}{\ddag} & 0.0016$\pm$0.00\textcolor{red}{\ddag} & 0.7219$\pm$0.00\textcolor{red}{\ddag} \\
                          & & & ✓ & 0.0135$\pm$0.00\textcolor{red}{\ddag}& 0.0016$\pm$0.00\textcolor{red}{\ddag} & 0.5659$\pm$0.00\textcolor{red}{\ddag} \\  \midrule

\multirow{3}{*}{LM}   & ✓ & ✓ & ✓ & 0.1001$\pm$0.04\textcolor{white}{\ddag} & 0.1937$\pm$0.08\textcolor{white}{\ddag} & 0.8503$\pm$0.04\textcolor{white}{\ddag} \\
                        & & ✓ & ✓ & -0.0016$\pm$0.02\textcolor{red}{\ddag} & 0.0074$\pm$0.02\textcolor{red}{\ddag} & 1.3347$\pm$0.06\textcolor{red}{\ddag} \\
                          & & & ✓ & -0.0039$\pm$0.00\textcolor{red}{\ddag} & 0.0025$\pm$0.00\textcolor{red}{\ddag} & 1.1030$\pm$0.00\textcolor{red}{\ddag} \\ \midrule
                            
\multirow{3}{*}{PB}   & ✓ & ✓ & ✓ & 0.1055$\pm$0.02\textcolor{white}{\ddag} & 0.1047$\pm$0.02\textcolor{white}{\ddag} & 0.1565$\pm$0.06\textcolor{white}{\ddag} \\
                        & & ✓ & ✓ & 0.0891$\pm$0.02\textcolor{red}{\ddag} & 0.0979$\pm$0.02\textcolor{red}{\ddag} & 1.6877$\pm$0.25\textcolor{red}{\ddag}\\
                          & & & ✓ & -0.0127$\pm$0.00\textcolor{red}{\ddag} & 0.0276$\pm$0.00\textcolor{red}{\ddag} & 0.9413$\pm$0.00\textcolor{red}{\ddag} \\ \midrule
                            
\multirow{3}{*}{RA}   & ✓ & ✓ & ✓ & 0.4033$\pm$0.08\textcolor{white}{\ddag} & 0.3444$\pm$0.02\textcolor{white}{\ddag} & 0.2574$\pm$0.08\textcolor{white}{\ddag} \\
                        & & ✓ & ✓ & 0.3673$\pm$0.01\textcolor{red}{\ddag} & 0.2911$\pm$0.02\textcolor{red}{\ddag} & 0.6584$\pm$0.02\textcolor{red}{\ddag} \\
                          & & & ✓ & 0.1758$\pm$0.00\textcolor{red}{\ddag} & 0.2554$\pm$0.00\textcolor{red}{\ddag} & 0.5323$\pm$0.00\textcolor{red}{\ddag} \\ \midrule
                            
\multirow{3}{*}{SE}   & ✓ & ✓ & ✓ & 0.7121$\pm$0.20\textcolor{white}{\ddag} & 0.7154$\pm$0.19\textcolor{white}{\ddag} & 0.5450$\pm$0.02\textcolor{white}{\ddag} \\
                        & & ✓ & ✓ & 0.4052$\pm$0.03\textcolor{red}{\ddag} & 0.4996$\pm$0.01\textcolor{red}{\ddag} & 0.7902$\pm$0.00\textcolor{red}{\ddag} \\
                          & & & ✓ & 0.0334$\pm$0.01\textcolor{red}{\ddag} & 0.0743$\pm$0.02\textcolor{red}{\ddag} & 1.6588$\pm$0.13\textcolor{red}{\ddag} \\ \midrule
                            
\multirow{3}{*}{WC}   & ✓ & ✓ & ✓ & 0.0447$\pm$0.00\textcolor{white}{\ddag} & 0.0419$\pm$0.01\textcolor{white}{\ddag} & 0.1707$\pm$0.03\textcolor{white}{\ddag} \\
                        & & ✓ & ✓ & 0.0295$\pm$0.01\textcolor{red}{\ddag} & 0.0074$\pm$0.00\textcolor{red}{\ddag} & 1.9827$\pm$0.35\textcolor{red}{\ddag} \\
                          & & & ✓ & -0.0106$\pm$0.00\textcolor{red}{\ddag} & 0.0052$\pm$0.00\textcolor{red}{\ddag} & 1.2073$\pm$0.00\textcolor{red}{\ddag} \\
                             
\bottomrule
\end{tabular}
\end{table}

A streaming data chunk generation algorithm named Two-Layer Random Sampling (TLRS) described in Algorithm~\ref{ATLRS} is designed to more realistically validate the performance of clustering methods on the ISDC problem. A selected dataset serves as the basis for generating data chunks. 
Through controlling the imbalance ratio and number of imbalanced clusters, TLRS can generate arbitrary-sized data chunks with $k^*$ imbalanced clusters to include various imbalance states of streaming data while preserving the original data distribution. The TLRS serves to enhance the diversity of experimental data, allowing a more convincing ISDC performance evaluation under various imbalanced scenarios.

The eight counterparts include the advanced automatic $k$-selection algorithms (CPCL \cite{Jia2014CPCL}, M3W \cite{m3w}), advanced algorithms designed for streaming data (StreamKM++ \cite{StreamKM++}, BIRCH \cite{BIRCH}), and state-of-the-art methods suitable for static imbalanced data (SMCL \cite{lu2019self}, IGMTT \cite{Cheung2019DenIGMTT}, DenSOINN \cite{Xu2019DenSIONN}, LDPI \cite{Tong_Wang_Liu}).
For StreamKM++ and BIRCH, the number of clusters needs to be specified in advance. In contrast, CPCL, SMCL, IGMTT, DenSOINN, LDPI, M3W and our method can adaptively determine the number of clusters without prior specification. Hyperparameters of our method are set as follows: the initial number of subnetworks is set at $T=15$, the number of objects $\rho$ for triggering Algorithm~\ref{Alg_IGCM} is set at $\rho=100$ recommended by \cite{GCSOM}, the number of nearest neighbors is set at $s=10$, and the learning rates $\epsilon_b$, $\epsilon_{\Omega}$, and $\alpha$ are set at 0.6, 0.02, and 0.005, respectively. The other parameters for all the compared methods are set following the recommendations in the source literature.

Three metrics, i.e., Adjusted Rand Index (ARI), Normalized Mutual Information (NMI), and Davies-Bouldin Index (DBI), are utilized for evaluation. ARI indicates the agreement in clustering that would be expected by random chance, which is a discriminative index with a range of $[-1, 1]$. NMI reflects the correlation between the clustering results and the given labels from the perspective of information theory, and its value range is $[0, 1]$. For both ARI and NMI, the larger their values, the better the clustering performance is. DBI is derived from the principles of internal consistency and density gap in cluster analysis. It evaluates the quality of clustering by measuring the distinction between adjacent clusters while also considering the tightness within clusters. A lower DBI value indicates better clustering, as it suggests that data objects within clusters are more compact and there is a higher degree of density gap between different clusters. The DBI range is $[0, \infty)$, where 0 represents a perfect clustering effect, i.e., clusters are very tight internally and completely separable from the others.  
The experiments are programmed using Python 3.11 and implemented using a workstation with 16GB RAM and 2.4GHz AMD R9 7940HX CPU.

\subsection{Efficiency Evaluation}

\begin{figure}[!t]
\centering
\begin{minipage}{1\linewidth}	
    \includegraphics[width=1\linewidth]{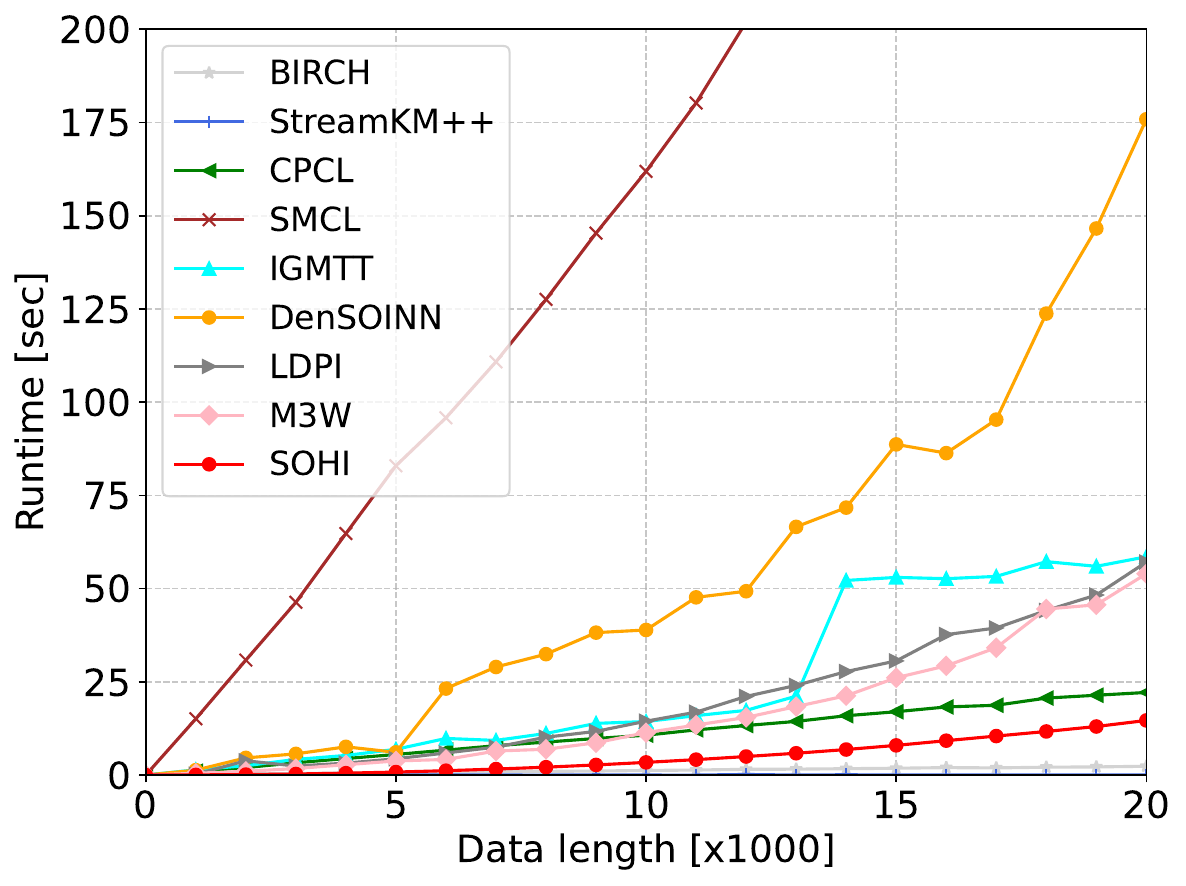}
 \end{minipage}	
 \caption{Effect of increasing data size on running time.}
\label{fig:running_time}
\end{figure}

The impact of data size on the running time of clustering methods is studied by plotting their running time with the increase of data size in Fig. \ref{fig:running_time}.  
It can be seen that the proposed SOHI has a similar running time as that of StreamKM++, BIRCH, and CPCL, thanks to low complexity, while the group of SMCL and DenSOINN is with heavy computational cost due to their polynomial time complexity.

\subsection{Clustering Accuracy Evaluation}

The clustering performance of different clustering methods is compared on all the datasets, where each dataset is generated with 10 chunks, and the average experimental results are reported. The best and second-best-performing methods on each dataset are highlighted in orange and gray, respectively. 

\begin{figure*}[!t]
\centering
\begin{minipage}{0.9\linewidth}	
    \includegraphics[width=1\linewidth]{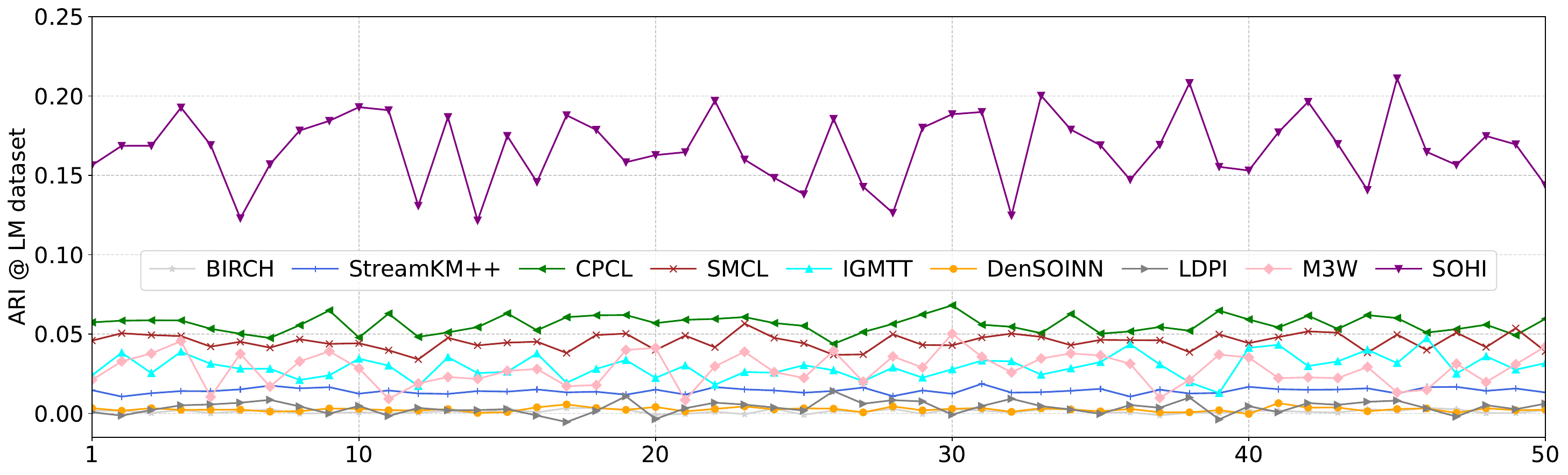}
 \end{minipage}	
 \caption{ARI performance on 50 streaming chunks of LM dataset.}
\label{fig:chunk50_ari}
\end{figure*}

\begin{figure*}[!t]
\centering
\begin{minipage}{0.9\linewidth}	
    \includegraphics[width=1\linewidth]{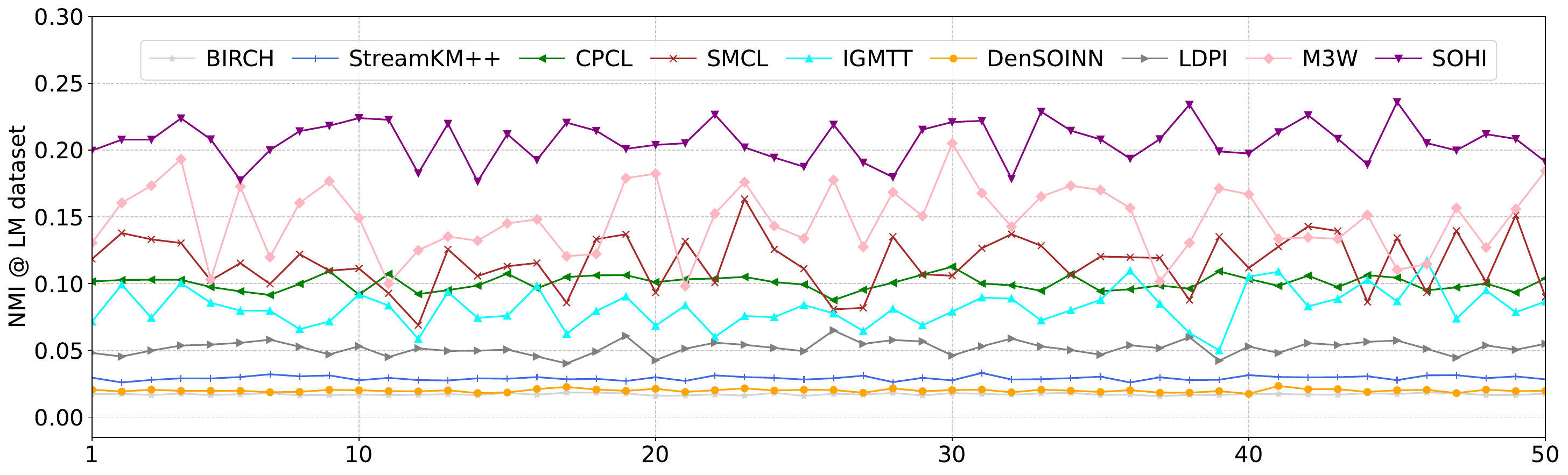}
 \end{minipage}	
 \caption{NMI performance on 50 streaming chunks of LM dataset.}
\label{fig:chunk50_nmi}
\end{figure*}

\begin{figure*}[!t]
\centering
\begin{minipage}{0.9\linewidth}	
    \includegraphics[width=1\linewidth]{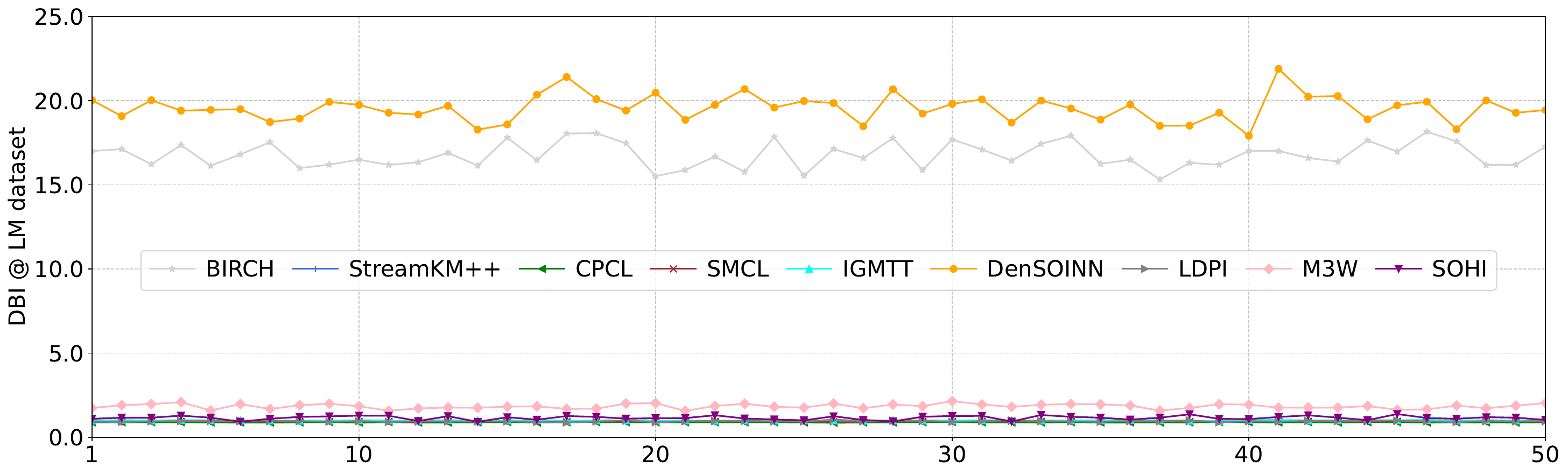}
 \end{minipage}	
 \caption{DBI performance on 50 streaming chunks of LM dataset.}
\label{fig:chunk50_dbi}
\end{figure*}

From Table~\ref{tb:result}, it can be observed that the proposed SOHI performs the best in general, winning or being the runner-up in 28 out of 33 comparisons. In comparison with the fast clustering methods, i.e., StreamKM++, BIRCH, and IGMTT, our SOHI demonstrates its superiority in accuracy, as it is competent in detecting imbalanced clusters, while the three fast methods do not adopt a mechanism to specially take the imbalanced issue into account. The results of LDPI on HF dataset and the results of M3W on CE dataset are not reported, because they wrongly group all the data objects into one cluster. LDPI's criteria for selecting initial subclusters is highly sensitive to the data distributions. It fails in clustering the HF dataset because a part of the objects is distributed with an extremely high density than the other parts. This results in only one subcluster being initialized by the LDPI. As for M3W, it fails on the CE dataset because the data objects are distributed in a sparse and relatively uniform way. Even the smallest number of neighbors suggested by M3W is still too large, leading to the merging of all the initialized cores to form a single cluster. In addition, since M3W tends to partition data objects into a larger number of smaller clusters, it is reasonable that M3W achieves superior clustering performance on AB dataset with many (i.e., $k^*=28$) clusters and performs well in terms of the internal DBI index. In general, SOHI demonstrates higher clustering accuracy in comparison with the fast algorithms for streaming data processing, while being extremely competitive in accuracy compared to the state-of-the-art algorithms proposed for static data clustering.

Then we evaluate the performance of different methods in adapting to consecutive imbalanced streaming data chunks. 50 chunks are generated using TLRS, and the ARI, NMI, and DBI performance of the methods per chunk (time-stamp) is shown in Figs.~\ref{fig:chunk50_ari} - \ref{fig:chunk50_dbi}, respectively. It can be observed that SOHI still significantly outperforms the other methods in general, which conforms with the observations of Table~\ref{tb:result}. Since the static data-oriented CPCL, M3W, SMCL, DenSOINN, and LDPI are also executable on streaming data chunks by treating each chunk as a static dataset, their ISDC performance is also reported. In summary, the proposed SOHI is superior in terms of clustering accuracy in handling imbalanced streaming data chunks with various imbalance ratios.

\subsection{Ablation Study}

To more specifically validate SOHI's effectiveness, we conduct ablation experiments to compare SOHI and its ablated variants. Since the merits of SOHI mainly stem from the initialization of Multiple Subnetworks (MS), the learning of SGM, and the HM module that obtains a proper number of clusters, the ablated variants are formed as follows: 1) To ablated MS, a single network is trained without MS initialization and subnetwork fusion; 2) To ablate SGM, the network's growth is restricted by preventing new neuron addition; 3) Since the effectiveness of HM is in the efficiency perspective, the HM module is treated as a module in this experiment. Because the ablated versions cannot handle streaming data, the ablation study is conducted on each whole static dataset. This is why the ablation study results in Table~\ref{tb:ablation} are not exactly the same as the chunk-wise results in Table~\ref{tb:result}. 

It can be observed that the ablation of any module of SOHI leads to a decrease in its clustering performance, indicating that each module contributes to achieving good clustering performance. More specifically, the MS ablation has a smaller impact, while the SGM ablation results in more significant accuracy differences. This is because even if MS is replaced by a single network, a considerable number of neurons can still represent the data distribution. However, for the SGM module, when it is restricted to grow, the limited number of neurons cannot finely describe the data distribution, thus severely influencing the following subnetwork fusion and cluster merging. In short, SOHI consistently produced the best clustering results across all ablation versions, confirming the effectiveness of the key components.

\definecolor{lgreen}{RGB}{144,238,144} 
\definecolor{mgreen}{RGB}{0,128,0} 
\definecolor{dgreen}{RGB}{0,100,0} 

\section{Conclusion}\label{section6}

An accurate and efficient ISDC method called SOHI is proposed. It adaptively trains growing neuron maps named SGM to achieve a topological representation of data distribution with rich local density information. The structure is proven to be: 1) efficient in adapting to new data distributions by incrementally updating its neurons, 2) effective in describing relatively small cluster distributions, and 3) efficient in providing retrieval information for microcluster merging. In the process of hierarchically merging microclusters to explore imbalanced clusters, the density distribution reflected by SGM is utilized to make a fine judgment on whether two microclusters should be merged. Such a process also guides the selection of the final appropriate number of clusters. To facilitate convincing experimental evaluation, we also propose a streaming data chunk generator that can simulate various extreme situations in real streaming data scenarios. Extensive experiments, including clustering accuracy and efficiency evaluation on streaming and static data, ablation studies, etc., have been conducted. By comparing with the state-of-the-art methods on various datasets, the proposed method is proven to be superior in both accuracy and efficiency for ISDC.

\section*{Acknowledgments}

The authors would like to thank Mr. Zhanpei Huang for providing the computational resources necessary to complete the efficiency evaluation experiments, Mr. Peilin Zhan for the insightful discussions regarding the learning mechanism of SGM, and Mr. Yongqi Xu and Mr. Yujian Lee for conducting the preliminary experiments in verifying the impact of the ISDC on clustering accuracy and efficiency.

\printbibliography  

\end{document}